\documentclass[10pt,twocolumn,letterpaper]{article}

\usepackage{cvpr}              
\usepackage[accsupp]{axessibility}

\usepackage{graphicx}
\usepackage{amsmath}
\usepackage{amssymb}
\usepackage{booktabs}

\usepackage{makecell}
\usepackage{array}
\usepackage{multirow,multicol}
\usepackage{tabulary}
\usepackage[dvipsnames]{xcolor}
\usepackage{soul}
\usepackage{xspace}\xspace
\usepackage{adjustbox}
\usepackage{array}
\usepackage{dblfloatfix}
\usepackage{transparent}
\usepackage{algorithm}
\usepackage{listings}
\usepackage{subcaption}
\usepackage[pagebackref,breaklinks,colorlinks]{hyperref}
\newcolumntype{x}[1]{>{\centering\arraybackslash}p{#1pt}}
\newcolumntype{y}[1]{>{\raggedright\arraybackslash}p{#1pt}}
\newcolumntype{z}[1]{>{\raggedleft\arraybackslash}p{#1pt}}
\newlength\savewidth\newcommand\shline{\noalign{\global\savewidth\arrayrulewidth
  \global\arrayrulewidth 1pt}\hline\noalign{\global\arrayrulewidth\savewidth}}

\newcommand{\green}[1]{\textcolor[RGB]{96,177,87}{#1}}
\newcommand{\blue}[1]{\textcolor[RGB]{30,144,255}{#1}}
\newcommand{\orange}[1]{\textcolor[RGB]{255,165,0}{#1}}

\newcommand{\fn}[1]{\footnotesize{#1}}

\newcommand{\gray}[1]{\textcolor{gray}{#1}}
\newcommand{\gbf}[1]{\green{\bf{\fn{(#1)}}}}

\usepackage[capitalize]{cleveref}
\crefname{section}{Sec.}{Secs.}
\Crefname{section}{Section}{Sections}
\Crefname{table}{Table}{Tables}
\crefname{table}{Tab.}{Tabs.}

\def\cvprPaperID{***} 
\def\confName{CVPR}
\def\confYear{2022}

\usepackage{graphicx}

\begin{document}

\title{DiRA: Discriminative, Restorative, and Adversarial Learning \\ for Self-supervised Medical Image Analysis}
\author{Fatemeh Haghighi$^{1}$\thanks{Equal contributors ordered alphabetically.}
\and
Mohammad Reza Hosseinzadeh Taher$^{1}$\footnotemark[1]
\and
Michael B. Gotway$^2$
\and
Jianming Liang$^1$\\
$^1$Arizona State University \thickspace \thickspace \thickspace \thickspace \thickspace $^2$Mayo Clinic\\
{\tt\small \{fhaghigh,mhossei2,jianming.liang\}@asu.edu} \thickspace \thickspace \thickspace \thickspace \thickspace {\tt\small Gotway.Michael@mayo.edu} 
}

\maketitle

\begin{abstract}
\noindent Discriminative learning, restorative learning, and adversarial learning have proven  beneficial for self-supervised learning schemes in computer vision and medical imaging. Existing efforts, however, omit their synergistic effects on each other in a \underline{ternary} setup, which, we envision, can significantly benefit deep semantic representation learning. To realize this vision, we have developed \textbf{DiRA}, the first framework that unites discriminative, restorative, and adversarial learning in a unified manner to collaboratively glean complementary visual information from unlabeled medical images for fine-grained semantic representation learning. Our extensive experiments demonstrate that DiRA (1) encourages collaborative learning among three learning ingredients, resulting in more generalizable representation across organs, diseases, and modalities; (2) outperforms fully supervised ImageNet models and increases robustness in small data regimes, reducing annotation cost across multiple medical imaging applications; (3) learns fine-grained semantic representation, facilitating  accurate lesion localization with only image-level annotation; and (4) enhances state-of-the-art restorative approaches, revealing that DiRA is a general mechanism for united representation learning. All code and pretrained models are available at \url{https://github.com/JLiangLab/DiRA}.
 
\end{abstract}

\section{Introduction}
\label{sec:intro}

\begin{figure}[t]
  \centering
  \includegraphics[width=0.7\linewidth]{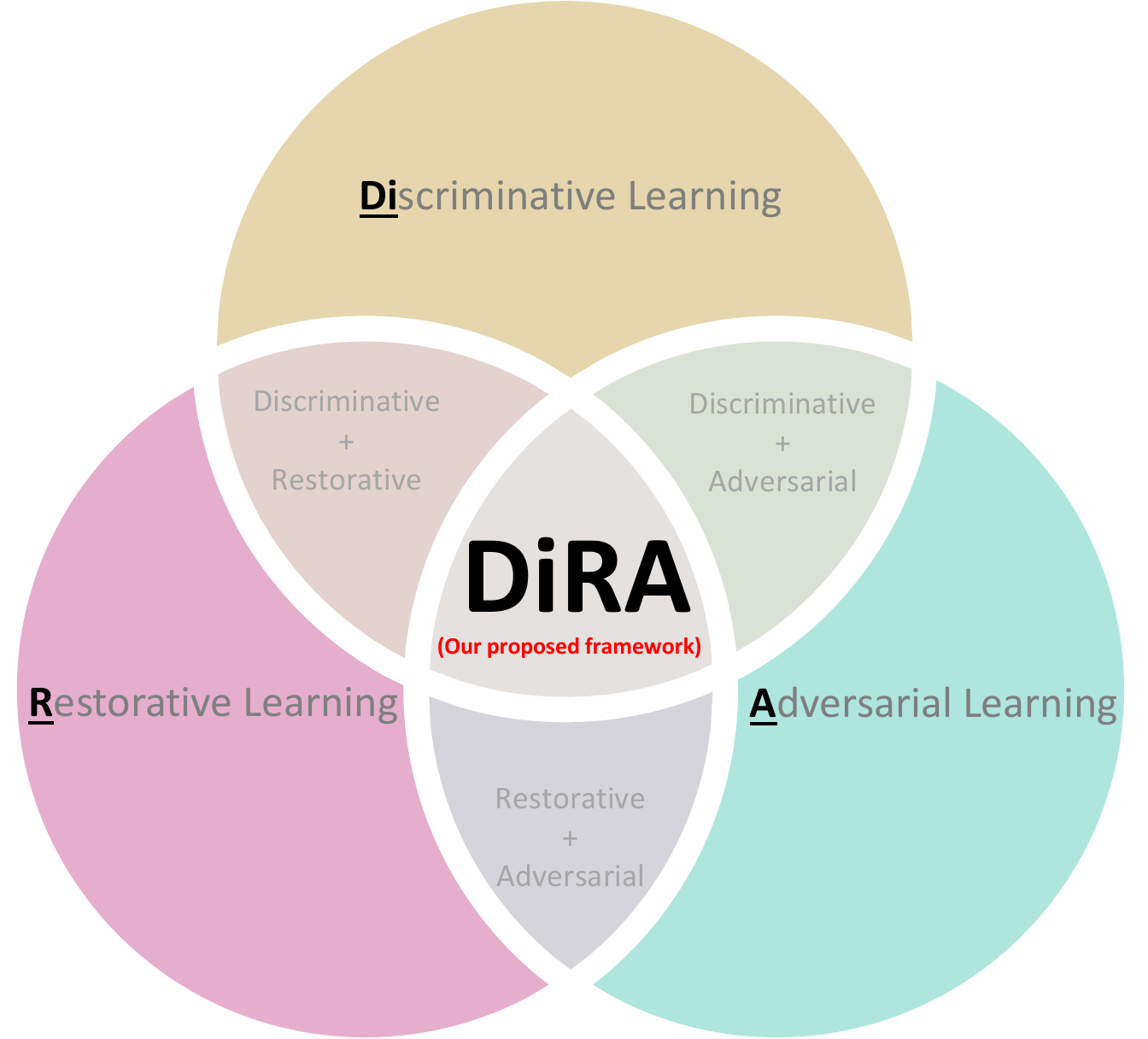}
   \caption{Despite the critical contributions of discriminative, restorative, and adversarial learning to SSL performance, yet no SSL method simultaneously employs all three learning ingredients. Our proposed DiRA, a novel SSL framework, unites discriminative, restorative, and adversarial learning in a unified manner to collaboratively glean complementary visual information from unlabeled data for fine-grained semantic representation learning.}   
   \label{fig:idea}
\end{figure}
Self-supervised learning (SSL) aims to learn generalizable representations without using any expert annotation. The representation learning approaches in the SSL paradigm can be categorized into three main groups: (1) \textit{discriminative learning}, which utilizes encoders to cluster instances of the same (pseudo) class and distinguish instances from different (pseudo) classes; (2) \textit{restorative learning}, which utilizes generative models to reconstruct original images from their distorted versions; and (3) \textit{adversarial learning}, which utilizes adversary models to enhance restorative learning.  In computer vision, discriminative SSL approaches, especially contrastive learning~\cite{Chen2020Simple,Chen2021Exploring,zbontar2021barlow,Grill2020Bootstrap,He2020Momentum,caron2021unsupervised,li2021prototypical,chen2020big,Ermolov2021Whitening,tian2020makes,He2020Momentum}, currently offer state-of-the-art (SOTA) performance, surpassing standard supervised ImageNet models in some tasks. In medical imaging, however, restorative SSL methods~\cite{ZHOU2021Models,chen2019self,haghighi2021transferable,Tao2020Revisiting,Zhou2021Preservational,haghighi2020learning} compared with discriminative approaches~\cite{azizi2021big,Zhou2020Comparing} presently reach a new height in performance. Naturally, we contemplate: \textit{What contributes to the popularity differences between discriminative and restorative methods in computer vision and in medical imaging?} Furthermore, from our extensive literature review, we have discovered that \textit{no} SSL method exploits all three learning components simultaneously; therefore, we ponder: \textit{Can discriminative, restorative, and adversarial learning be seamlessly integrated into a single framework to foster collaborative learning for deep semantic representation, yielding more powerful models for a broad range of applications?} In seeking answers to the two questions, we have gained the following insights.

\begin{figure}[t]
  \centering
  \includegraphics[width=0.9\linewidth]{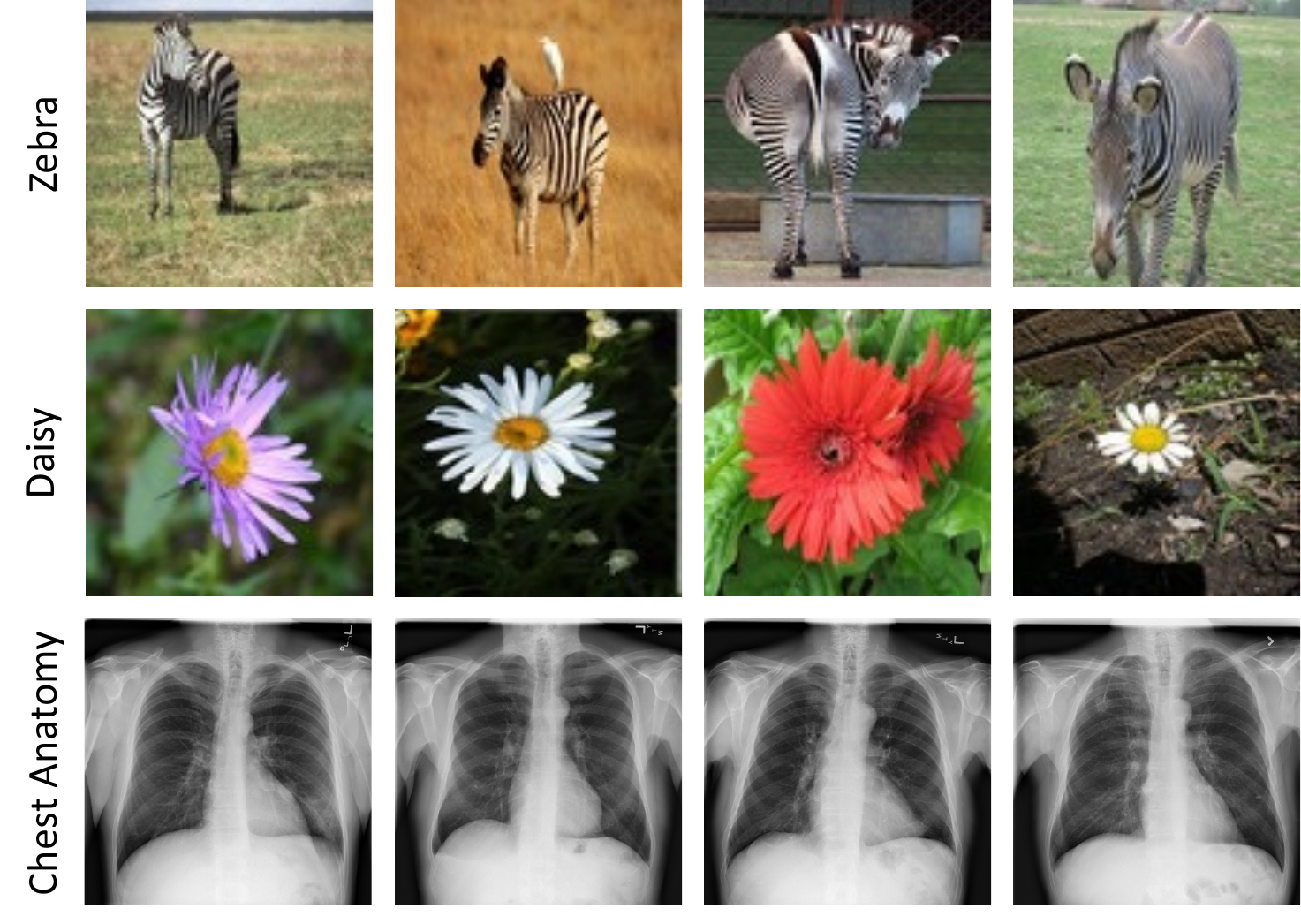}
   \caption{Photographic images typically have large foreground objects with apparent discriminative parts, whereas medical images contain consistent anatomical structures with semantic information dispersed over the entire images. As a result, recognition tasks in photographic images are mainly based on high-level features, while medical tasks demand holistic fine-grained discriminative features captured throughout images.}   
   \label{fig:natural_vs_medical}
\end{figure}

Computer vision and medical imaging tasks embrace the spirit of evil in opposite ways, originating from the \textit{marked} differences between photographic and medical images. Photographic images, particularly those in ImageNet, have large foreground objects with apparent discriminative parts, residing in varying backgrounds (e.g.\ zebra and daisy images in~\cref{fig:natural_vs_medical}).  Thus, object recognition tasks in photographic images are primarily based on high-level features captured from discriminative regions. In contrast, medical images generated from a particular imaging protocol exhibit consistent anatomical structures (e.g, chest anatomy in ~\cref{fig:natural_vs_medical}), with clinically relevant information dispersed over the entire image~\cite{haghighi2021transferable}. In particular, high-level structural information, i.e., anatomical structures and their relative spatial orientations, are essential for the identification of normal anatomy and various disorders. Importantly, medical tasks require much stronger attention to fine-grained details within images as identifying diseases, delineating organs, and isolating lesions rely on subtle, local variations in texture~\cite{Taher2021Systematic}.  Therefore, recognition tasks in medical images desire complementary high-level and fine-grained discriminative features captured throughout images.

According to our systematical analysis, we have gained the following understandings:  (1) discriminative learning excels in capturing high-level (global) discriminative features, (2) restorative learning is good at conserving fine-grained details embedded in local image regions, and (3) adversarial learning consolidates restoration by conserving more fine-grained details. Putting these understandings and fundamental differences between photographic and medical images together would explain why restorative learning is preferred in medical imaging while discriminative learning is preferred in computer vision. More importantly, we have acquired a new and intriguing insight into \textit{trio} of discriminative, restorative, and adversarial learning to excavate effective features required for medical recognition tasks---not only high-level anatomical representations but also fine-grained discriminative cues embedded in the local parts of medical images.

Based on the insights above, we have designed a novel self-supervised learning framework, called DiRA, by uniting \textbf{di}scriminative learning, \textbf{r}estorative learning, and \textbf{a}dversarial learning in a unified manner to glean complementary visual information from unlabeled medical images. Our extensive experiments demonstrate that (1) DiRA encourages collaborative learning among three learning components, resulting in more generalizable representation across organs, diseases, and modalities (see \cref{fig:w_wo_red}); (2) DiRA outperforms fully supervised ImageNet models and increases robustness in small data regimes, thereby reducing annotation cost in medical imaging (\cref{tab:annotation} and \cref{tab:2d_sota}); (3) DiRA learns fine-grained representations, facilitating more accurate lesion localization with only image-level annotations (\cref{fig:CAM}); and (4) DiRA enhances SOTA restorative approaches, showing that DiRA is a general framework for united representation learning (\cref{tab:sota_3D}).

In summary, we make the following contributions:
\begin{itemize}
    \item The insights that we have gained into the \textit{synergy} of discriminative, restorative, and adversarial learning in a \underline{ternary} setup, realizing a new paradigm of collaborative learning for SSL.
    
    \item The first self-supervised learning framework that seamlessly unites discriminative, restorative, and adversarial learning in a unified manner, setting a new SOTA for SSL in medical imaging.
    
    \item A thorough and insightful set of experiments that demonstrate not only DiRA's generalizability but also its potential to take a fundamental step towards developing \textit{universal} representations for medical imaging.
\end{itemize}

\section{Related works}
\noindent\textbf{Discriminative self-supervised learning.} Discriminative methods can be divided into \textit{class-level} and \textit{instance-level} discrimination. \textit{Class-level} discrimination methods~\cite{gidaris2018unsupervised,Zhan2020Online, caron2018deep,caron2021unsupervised,noroozi2016unsupervised,doersch2015Unsupervised} group images based on certain criteria, assign a pseudo label to each group, and train a model to discriminate the images based on their pseudo labels, such as rotation degrees~\cite{gidaris2018unsupervised} and cluster assignments~\cite{Zhan2020Online, caron2018deep,caron2021unsupervised}. On the other hand, \textit{instance-level} discrimination methods~\cite{Wu2018Unsupervised,Chen2020Simple,Chen2021Exploring,zbontar2021barlow,Grill2020Bootstrap,He2020Momentum,caron2021unsupervised,li2021prototypical,chen2020big,Ermolov2021Whitening,tian2020makes,Ye2019Unsupervised} treat each image as a distinct class, and maximize the similarity of representations derived from different views of the same image, seeking to learn transformation invariant representations. \textit{Instance-level} discriminative learning has been investigated in various forms, including contrastive learning~\cite{He2020Momentum,Chen2020Simple,Wu2018insdis,chen2020improved}, asymmetric networks~\cite{Chen2021Exploring,Grill2020Bootstrap}, and redundancy reduction~\cite{zbontar2021barlow,Ermolov2021Whitening}. However, both \textit{class-level} and \textit{instance-level} approaches in discriminative learning have shown failures in tasks that require finer-grained features~\cite{Wang2021Dense,Xie2021Propagate,Xie2021DetCo}. Our DiRA addresses this limitation by incorporating restorative and adversarial learning, which not only improves discriminative learning but also yields fine-grained representations required for medical imaging tasks.

\smallskip
\noindent\textbf{Restorative and adversarial self-supervised learning.}  The key objective for a restorative method is to faithfully reconstruct the distribution of data~\cite{Wu2018Unsupervised,Parmar2021Dual}. 
    In the SSL context, multiple pretext tasks are formulated to reconstruct the perturbed images using generative models~\cite{pathak2016context,Vincent2008Extracting,larsson2017colorproxy}. The advance of GANs~\cite{NIPS2014Goodfellow} has led to a new line of research in unsupervised learning, using adversarial learning to generate transferable representations~\cite{Dumoulin2017Adversarially,Donahue2019Large}. While recent works~\cite{Chen2020Generative,Donahue2019Large} have demonstrated impressive results by employing large-scale generative models, it remains unclear to what extent generative models can encapsulate high-level structures. Our DiRA alleviates this limitation by bringing the advantages of discriminative learning into generative models. Through discriminating image samples, generative models are encouraged to capture global discriminative representations rather than superficial representations, leading to a more pronounced embedding space.

\smallskip
\noindent\textbf{Self-supervised learning in medical imaging.} 
Due to the lack of large annotated datasets, SSL created substantial interest in medical imaging. Motivated by the success in computer vision, recent discriminative methods concentrate on instance-level discrimination.  A comprehensive benchmarking study in~\cite{Taher2021Systematic} evaluated the efficacy of existing instance discrimination methods pretrained on ImageNet for diverse medical tasks. Several other works adjusted contrastive-based methods on medical images~\cite{azizi2021big,Chaitanya2020Contrastive,Zhou2020Comparing}. A large body of work, on the other hand, focuses on restorative approaches, which can be categorized into restorative  only~\cite{chen2019self,ZHOU2021Models}, restorative  and adversarial~\cite{Tao2020Revisiting}, and  discriminative  and restorative~\cite{haghighi2021transferable,Taher2022CAiD,Zhou2021Preservational}. Among these groups, the most recent study on TransVW~\cite{haghighi2021transferable,haghighi2020learning} demonstrated superiority by combining discriminative and restorative components into a single SSL framework. DiRA distinguishes itself from all previous works by demonstrating two key advances: (1) employing discriminative, restorative, and adversarial learning simultaneously in a unified framework; and (2) providing a general representation learning framework that is compatible with existing discriminative and restorative methods, regardless of their objective functions. 

\begin{figure}[t]
\centering
\includegraphics[width=0.85\linewidth]{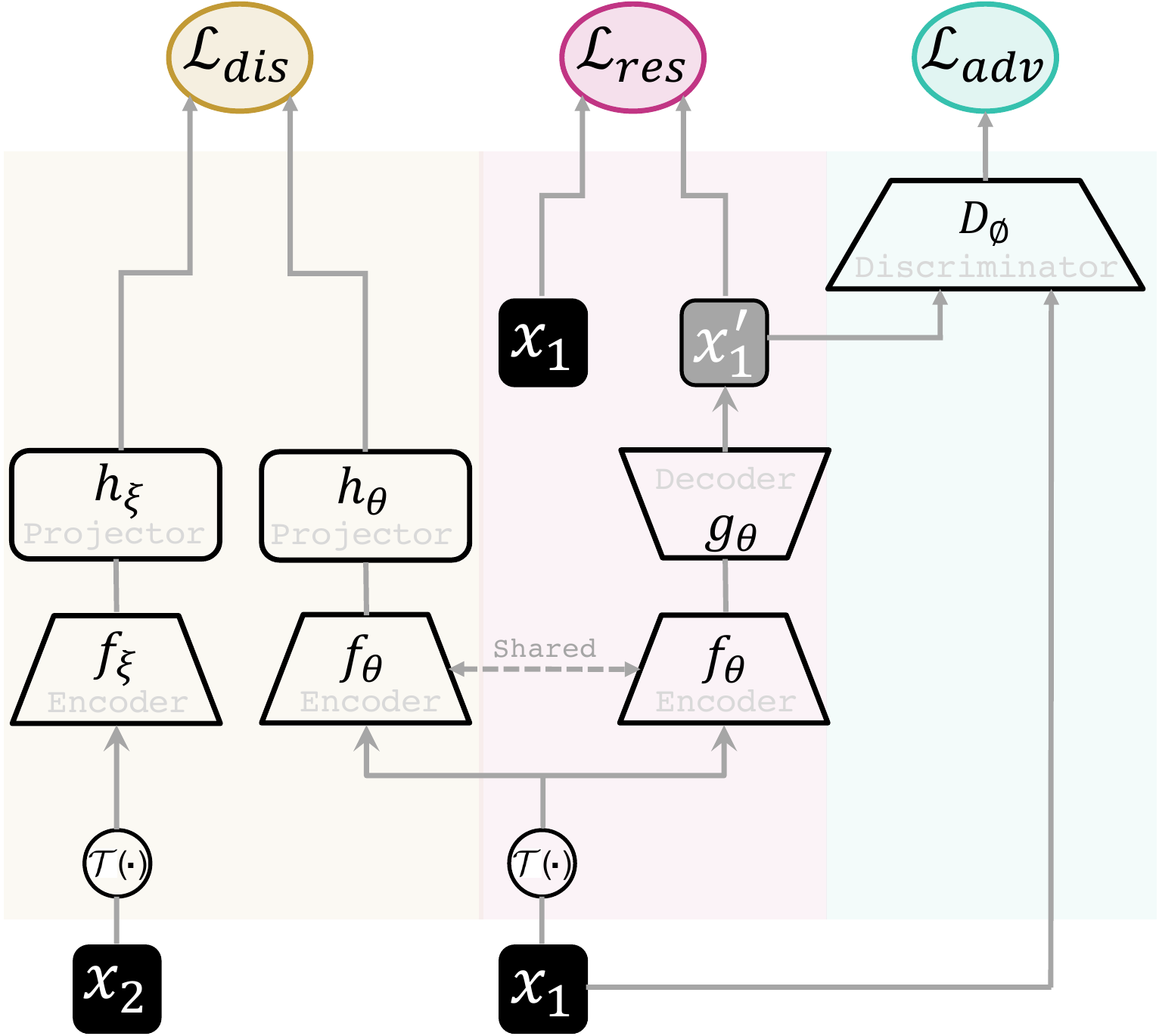}
\caption{\textbf{Our proposed framework.} DiRA consists of three learning components: discriminative, restorative, and adversary.  Given two input patches $x_1$ and $x_2$, we perturb them with $\mathcal{T}(.)$ and provide them as input to discrimination and restoration branches. The discrimination branch consists of encoders $f_\theta$ and $f_\xi$, and projectors $h_\theta$ and $h_\xi$, and maximizes the agreement between (high-level) embedding vectors of samples from the same (pseudo) class. The restoration branch consists of encoder $f_\theta$ and decoder $g_\theta$, and maximizes the (pixel-level) agreement between original sample $x_1$ and  restored $x'_1$. Adversarial discriminator $D_\phi$ contrasts the original samples with the restored ones, reinforcing the restoration to preserve more fine-grained details.}
\label{fig:method}
\end{figure}

\section{DiRA framework}
As shown in~\cref{fig:method}, DiRA is a SSL framework comprised of three key components: (1) Discrimination (Di) that aims to learn high-level discriminative representations,  (2) Restoration  (R) that aims to enforce the model to conserve fine-grained information about the image by focusing on more localized visual patterns, and (3) Adversary (A) that aims to further improve feature learning through the restoration component. By integrating these  components into a \textit{unified} framework, DiRA captures comprehensive information from images, providing more powerful representations for various downstream tasks. In the following, we first introduce each component by abstracting a common paradigm and then describe the joint training loss.

\subsection{Discriminative learning}
 Discriminative learning can be thought of as training an encoder to maximize agreement between instances of the same (pseudo) class in the latent space via a discriminative loss. As illustrated in \cref{fig:method}, the discriminator branch is comprised of two twin backbone networks $f_\theta$ and $f_\xi$, and projection heads $h_\theta$ and $h_\xi$. $f_\theta$ is a regular encoder, while $f_\xi$ can be a momentum encoder \cite{He2020Momentum,Grill2020Bootstrap} or share weights with $f_\theta$~\cite{zbontar2021barlow,Chen2021Exploring,haghighi2021transferable}. Given two patches $x_1$ and $x_2$, which are cropped from the same image or different images, we first apply an augmentation function $\mathcal{T}(.)$ on them. The two augmented patches are then processed by $f_\theta$ and $f_\xi$ to generate latent features $y_1=f_\theta(\mathcal{T}(x_1))$ and $y_2=f_\xi(\mathcal{T}(x_2))$.
  The projection heads $h_\theta$ and $h_\xi$ projects the latent features to a unit sphere and output projections $z_1=h_\theta(y_1)$ and $z_2=h_\xi(y_2)$. The discriminator's objective is to maximize the similarity between the embedding vectors obtained from two samples of the same (pseudo) class:  
\begin{equation}
  \mathcal{L}_{dis} = \ell( z_1,z_2)
  \label{eq:loss_D}
\end{equation}
where $\ell(z_1,z_2)$ is the similarity/distance function that measures compatibility between $z_1$ and $z_2$. 
DiRA is a general framework that allows various choices of discrimination tasks without any constraint. As such, the  declaration of class might range from considering every single image as a class (instance discrimination) to clustering images based on a similarity metric (cluster discrimination). Accordingly, $x_1$ and $x_2$ can be two views of the same image or two samples from the same cluster. Based on the nature of the discrimination task, the instantiation of $\mathcal{L}_{dis}$ can be cross-entropy~\cite{haghighi2021transferable,gidaris2018unsupervised,noroozi2016unsupervised,Zhuang2019Self}, contrastive~\cite{He2020Momentum,Chen2020Simple,azizi2021big,caron2021unsupervised}, redundancy reduction~\cite{zbontar2021barlow,Ermolov2021Whitening}, etc.

\subsection{Restorative learning}
Our restorative learning branch aims to enhance discrimination learning by leveraging fine-grained visual information.  As shown in \cref{fig:method}, the restoration branch is comprise of an encoder $f_\theta$ and decoder $g_\theta$, where encoder $f_\theta$ is shared with the discrimination branch. 
Given the input sample $x_1$ distorted by $\mathcal{T}$, the $f_\theta$ and $g_\theta$ aims to map the distorted sample back to the original one, i.e., $f_\theta,g_\theta:(x,\mathcal{T})\mapsto x$. 
$f_\theta$ and $g_\theta$ are trained by minimizing the distance between the original sample and the restored one at pixel-level:
\begin{equation}
  \mathcal{L}_{res} = \mathbb{E}_x \; dist(x_1, x_1') 
  \label{eq:loss_res}
\end{equation}
where $x'_1 = g_\theta(f_\theta(\mathcal{T}(x_1)))$ denotes the restored image. $dist(x_1, x_1')$ presents the distance function that measures similarity between $x_1$ and $x_1'$, such as $L_1$ or $L_2$.

\subsection{Adversarial learning}
Adversarial learning aims to reinforce $f_\theta$ by measuring how realistic the restored images are.
As such, the adversarial discriminator $D_\phi$ is formulated to distinguish (discriminate) the set of training images from the set of synthesized images, guiding encoder $f_\theta$ to capture more informative features from images so that  $g_\theta$ can reproduce the original images effectively. Therefore, the encoder $f_\theta$ and decoder $g_\theta$ play a minimax game with adversarial discriminator $D_\phi$, and are optimized jointly with an adversarial loss~\cite{CaoAuto2020,Parmar2021Dual}: 
\begin{equation}
\mathcal{L}_{adv} = \mathbb{E}_x \;  [log \; D_\phi(x_1)] \; + \mathbb{E}_x \; [log(1 -  D_\phi(x'_1))]
\end{equation}

\subsection{Joint training}
Finally, the combined objective for the proposed DiRA framework becomes:
\begin{equation}
  \mathcal{L} = \lambda_{dis}*\mathcal{L}_{dis} + \lambda_{res} * \mathcal{L}_{res} + \lambda_{adv} * \mathcal{L}_{adv} 
  \label{eq:loss_gan}
\end{equation}
where  $\lambda_{dis}$, $\lambda_{res}$, and $\lambda_{adv}$ are multiplication factors that determine the relative importance of different losses. Through our unified training scheme, DiRA learns a representation that preserves fine-grained details within the samples while being discriminative among the image classes. In particular, the formulation of $\mathcal{L}_{dis}$ encourages the model to capture high-level discriminative features. Moreover, $\mathcal{L}_{res}$ forces the model to encode fine-grained information from the images by focusing on pixel-level visual patterns. This  results in more descriptive  feature embeddings that elevate the discrimination task. Finally, $\mathcal{L}_{adv}$  elevates restoration based learning through capturing more informative features. 

\begin{figure*}[t]
  \centering
  \includegraphics[width=0.9\linewidth]{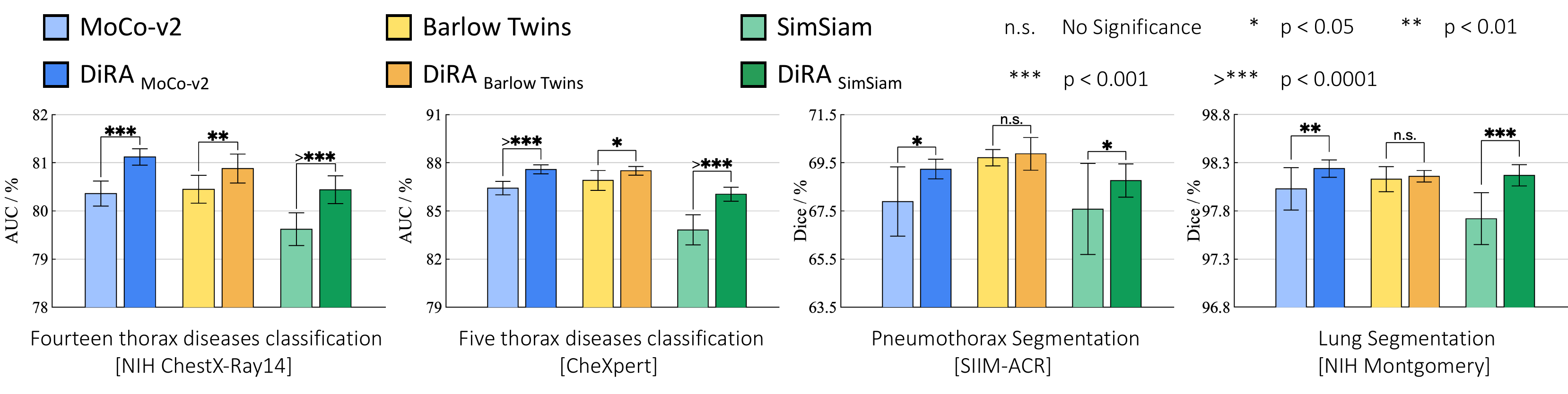}
   \caption{\textbf{Comparison with  discriminative self-supervised methods:} We apply our DiRA to three representative SOTA self-supervised methods with different discrimination objectives: MoCo-v2~\cite{chen2020improved}, Barlow Twins~\cite{zbontar2021barlow}, and SimSiam~\cite{Chen2021Exploring}. DiRA empowers discriminative methods to capture more fine-grained representations, yielding significant ($p < 0.05$) performance gains on four downstream tasks.
  }
   \label{fig:w_wo_red}
\end{figure*}

\section{Implementations details}
\subsection{Pretraining protocol}
\label{sec:pretraining_setup}
DiRA is a general framework that is compatible with existing  self-supervised discriminative and restorative methods, regardless of their objective functions. To assess the effectiveness of our  framework, we adapt recent SOTA 2D and 3D self-supervised methods into  DiRA, as described in the following. The pretrained models with DiRA are identified as DiRA subscripted by the original method name. 

\smallskip
\noindent\textbf{2D image pretraining settings.} We apply DiRA to MoCo-v2~\cite{chen2020improved}, Barlow Twins~\cite{zbontar2021barlow}, and SimSiam~\cite{Chen2021Exploring} for 2D image self-supervised learning. All DiRA models are pretrained from scratch on the training set of ChestX-ray14~\cite{wang2017chestx} dataset. For each of these three discrimination tasks~\cite{chen2020improved,zbontar2021barlow,Chen2021Exploring}, we follow the original methods in the formulation of $\mathcal{L}_{dis}$, projection head architecture, and hyper-parameters settings. Furthermore, we  optimize the encoder and decoder networks $f_\theta$ and $g_\theta$ following the optimization setups  in~\cite{chen2020improved,zbontar2021barlow,Chen2021Exploring}. For all methods, we employ a 2D U-Net~\cite{Ronneberger2015Unet} with a standard ResNet-50~\cite{he2016deep} backbone as the $f_\theta$ and $g_\theta$. We adopt mean square error (MSE) as the $\mathcal{L}_{res}$.
The adversarial discriminator network $D_\phi$ consists of four convolutional layers with the kernel size of 3$\times$3~\cite{pathak2016context}, trained using the Adam optimizer with a learning rate of 2e-4 and $(\beta_1,\beta_2) = (0.5, 0.999)$. We use batch size 256 distributed across 4 Nvidia V100 GPUs. $\lambda_{res}, \lambda_{adv}, \lambda_{dis}$ are empirically set to 10, 0.001, and 1, respectively. Input images are first randomly cropped and resized to 224$\times$224; the image augmentation function $\mathcal{T}(.)$ includes random horizontal flipping, color jittering, and Gaussian blurring. Additionally, we apply cutout~\cite{devries2017improved,pathak2016context} and shuffling~\cite{chen2019self} to make the restoration task more challenging. More details are provided in the Appendix.


\smallskip
\noindent\textbf{3D volume pretraining settings.} We apply DiRA to TransVW~\cite{haghighi2021transferable}, the SOTA method for 3D self-supervised learning in medical imaging. We adapt TransVW in DiRA by adding an adversarial discriminator $D_\phi$ into its training scheme. For fair comparisons, we follow the publicly available TransVW code for setting instance discrimination and restoration tasks. Moreover, similar to publicly released TransVW, DiRA models are pretrained from scratch using 623 chest CT scans in the LUNA~\cite{setio2017validation} dataset. We use 3D U-Net~\cite{Ellis2021Trialing} as the encoder-decoder network and a classification head including fully-connected layers. The adversarial discriminator $D_\phi$ includes four convolutional blocks with the kernel size 3$\times$3 $\times$3. $\lambda_{res}, \lambda_{adv}, \lambda_{dis}$ are empirically set to 100, 1, and 1, respectively. $f_\theta$, $g_\theta$, and  $D_\phi$, are optimized for 200 epochs using Adam with a learning rate of 1e-3 and batch size of 8. More  details are provided in the Appendix.

\subsection{Transfer learning protocol}
\label{sec:finetuning_setup}
\noindent\textbf{Target tasks and datasets.} We evaluate the effectiveness of DiRA's representations in transfer learning to a diverse suite of 9 common but challenging 2D and 3D medical imaging tasks, including: ChestX-ray14, CheXPert~\cite{irvin2019chexpert}, SIIM-ACR~\cite{PNEchallenge}, and NIH Montgomery~\cite{Jaeger2014Tow} for 2D models, and  LUNA, PE-CAD~\cite{tajbakhsh2015computer}, LIDC-IDRI~\cite{armato2011lung}, LiTS~\cite{bilic2019liver}, and BraTS~\cite{bakas2018identifying} for 3D models (see Appendix for dataset details). 
These tasks encompass various label structures (multi-label classification and pixel-level segmentation), diseases (brain tumors and thoracic diseases, such as lung nodules, pulmonary emboli, and pneumothorax), organs (lung, liver, brain), and modalities (X-ray, CT, MRI). Moreover, these tasks contain many hallmark challenges encountered  when working with medical images, such as imbalanced classes, limited data, and small-scanning areas for the pathology of interest~\cite{Taher2021Systematic,azizi2021big}. We use the official data split of these datasets when available; otherwise, we randomly divide the data into 80\%/20\% for training/testing.

\medskip
\noindent\textbf{Fine-tuning settings.} We transfer the pretrained (1)  encoder ($f_\theta$) to the classification tasks, and (2) encoder and decoder ($f_\theta$ and $g_\theta$) to segmentation tasks. We  evaluated the generalization of DiRA representations by fine-tuning all the parameters of downstream models. We use the AUC (area under the ROC curve), and the IoU (Intersection over Union) and Dice coefficient for evaluating classification and segmentation performances, respectively. Following~\cite{Taher2021Systematic}, we strive to optimize each downstream task with the best performing hyperparameters (details in Appendix). We employ the early-stop mechanism using 10\% of the training data as the validation set to avoid over-fitting. We run each method ten times on each downstream task and report the average, standard deviation, and statistical analysis based on an independent two-sample \textit{t}-test.

\begin{table*}[]
\centering
\scalebox{0.76}{
\begin{tabular}{l|l l l |l l l |l l l}
\shline
\multirow{3}{*}{Method} & \multicolumn{3}{c|}{ChestX-ray14 [AUC (\%)]} & \multicolumn{3}{c|}{CheXpert [AUC (\%)]} &
\multicolumn{3}{c}{Montgomery [Dice (\%)]}\\ 
\cline{2-4} \cline{5-7} \cline{8-10}
& \multicolumn{3}{c|}{Label fraction} & \multicolumn{3}{c|}{Label fraction} &
\multicolumn{3}{c}{Label fraction}\\
& 1\% & 25\% & 50\% & 1\% & 25\% & 50\% & 1\% & 25\% & 50\%\\ \shline
MoCo-v2~\cite{chen2020improved}& 52.99 & 74.89 & 76.71 &76.87 &81.70	&83.23 &63.69 &		96.44 &	97.60 \\
DiRA$_{\text{MoCo-v2}}$ & \bf{59.39} \gbf{$\uparrow$ 6.4} & \bf{77.55} \gbf{$\uparrow$ 2.6} &\bf{78.74} \gbf{$\uparrow$ 2.0}& \bf{78.43} \gbf{$\uparrow$ 1.5}& \bf{87.12} \gbf{$\uparrow$ 5.4}& \bf{87.31} \gbf{$\uparrow$ 4.0}& \bf{72.53} \gbf{$\uparrow$ 8.8} &	\bf{97.06} \gbf{$\uparrow$ 0.62} &	\bf{98.14} \gbf{$\uparrow$ 0.5} \\\hline

Barlow Twins~\cite{zbontar2021barlow} &62.43&76.23& 77.59& 82.85	&	83.74&	84.66&86.79	&	97.49 &	97.68 \\
DiRA$_{\text{Barlow Twins}}$ & \bf{62.51} \gbf{$\uparrow$ 0.08}&\bf{77.18} \gbf{$\uparrow$ 0.9}& \bf{78.46} \gbf{$\uparrow$ 0.8} &
\bf{83.12} \gbf{$\uparrow$ 0.2}	&	\bf{84.20} \gbf{$\uparrow$ 0.4}&	\bf{85.32} \gbf{$\uparrow$ 0.6} & \bf{87.25} \gbf{$\uparrow$ 0.4}	&	\bf{97.62} \gbf{$\uparrow$ 0.1} &	\bf{98.15} \gbf{$\uparrow$ 0.4}\\ \hline
SimSiam~\cite{Chen2021Exploring} & 51.07 & 73.05& 75.20& 65.39&80.05&81.46&48.20 &		94.86&	97.21 \\
DiRA$_{\text{SimSiam}}$ & \bf{53.42} \gbf{$\uparrow$ 2.3} & \bf{74.38} \gbf{$\uparrow$ 1.3}& \bf{76.43} \gbf{$\uparrow$ 1.2}&\bf{70.46} \gbf{$\uparrow$ 5.0}	&\bf{81.03} \gbf{$\uparrow$ 1.0}&\bf{82.70} \gbf{$\uparrow$ 1.2}&\bf{61.86} \gbf{$\uparrow$ 13.6}&	\bf{96.61} \gbf{$\uparrow$ 1.7}&	\bf{97.91} \gbf{$\uparrow$ 0.7}\\ \shline
\end{tabular}}
\caption{\textbf{Transfer learning under different downstream label fractions:} DiRA models combat overfitting in low data regimes and provide stronger representations for downstream tasks with limited annotated data.  For each downstream task, we report the average performance over multiple runs. \gbf{$\uparrow$} shows the improvement of DiRA models compared with the underlying discriminative method.
}
\label{tab:annotation}
\end{table*}

\begin{figure*}[t]
  \centering
  \includegraphics[width=0.9\linewidth]{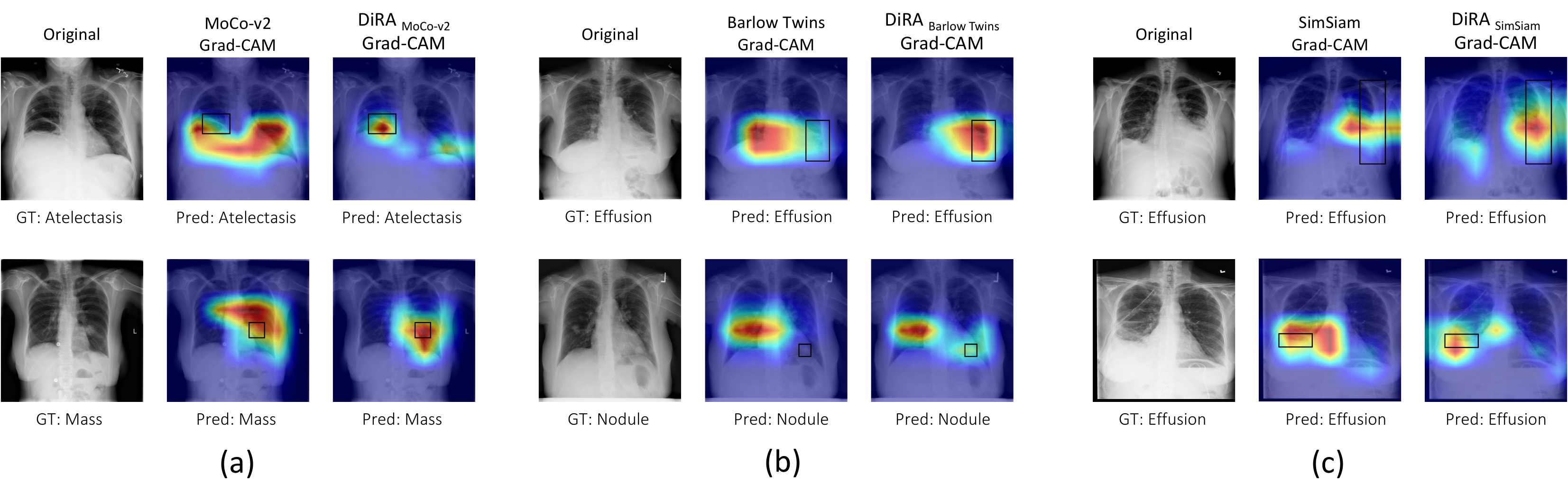}
   \caption{\textbf{Visualization of Grad-CAM heatmaps} for (a) MoCo-v2 vs. DiRA$_{\text{MoCo-v2}}$, (b) Barlow Twins vs. DiRA$_{\text{Barlow Twins}}$, and (c) SimSiam vs. DiRA$_{\text{SimSiam}}$. Ground truth bounding box annotations are shown in black. Training with DiRA leads to improvements in weakly-supervised disease localization. While both DiRA and underlying models predict the correct disease label on the test images, DiRA
models capture the diseased locations more precisely than the baselines which attune to larger regions of the image (e.g. (c), second row) or provide inaccurate localization with no overlap with the ground truth
(e.g. (b), second row).}
   \label{fig:CAM}
\end{figure*}

\section{Results}
We conduct extensive experiments to better understand not only the properties of our framework but also its generalizability across 9 downstream tasks. 
Through the following groups of experiments, we establish that DiRA  (1) enriches existing discriminative approaches, capturing a more diverse visual representation that generalizes better to different tasks; (2) addresses the annotation scarcity challenge in medical imaging, providing an annotation-efficient solution for medical imaging; (3) learns fine-grained features, facilitating more accurate lesion localization with only image-level annotation; and (4) improves SOTA restorative approaches, demonstrating that DiRA is a general framework for united representation learning. 

\subsection{DiRA enriches discriminative learning}
\noindent\textbf{Experimental setup:} To study the flexibility and efficacy of our proposed self-supervised framework, we apply DiRA to three recent SOTA self-supervised methods with diverse discrimination objectives: MoCo-v2, Barlow Twins, and SimSiam.  To evaluate the quality of our learned representations and ascertain the generality of our findings, we follow~\cite{Taher2021Systematic} and consider a broader range of four target tasks, covering classification (ChestX-Ray14 and CheXpert) and segmentation (SIIM-ACR and Montgomery).

\smallskip
\noindent\textbf{Results:} As seen in \cref{fig:w_wo_red}, utilizing our DiRA framework consistently enhances its underlying discriminative method across all tasks (1) ChestX-ray14, (2) CheXpert, (3) SIIM-ACR, and (4) NIH Montgomery. Compared with the original methods, DiRA$_{\text{MoCo-v2}}$ showed increased performance by 0.76\%, 1.17\%, 1.35\%, and 0.21\%, respectively; Similarly, DiRA$_{\text{Barlow Twins}}$ showed increased performance by 0.43\%, 0.60\%, 0.16\%, and 0.03\%. Finally, DiRA$_{\text{SimSiam}}$ showed increased performance by 0.82\%, 2.22\%, 1.18\%, and 0.45\%. These results imply that DiRA is a comprehensive representation learning framework that encourages existing self-supervised instance discriminative approaches to retain more fine-grained information from images, enriching their visual representation and allowing them to generalize to different medical tasks more effectively.

\subsection{DiRA improves robustness to small data regimes}
\noindent\textbf{Experimental setup:} We investigate the robustness of representations learned with  DiRA in  small data regimes to determine if the learned representation can serve as a proper foundation for fine-tuning. We randomly select 1\%, 25\%, and 50\% of training data  from ChestX-ray14, CheXpert, and Montgomery, and fine-tune the self-supervised pretrained models on these training-data subsets. 

\begin{table*}[]
\centering
\scalebox{0.8}{
\begin{tabular}{l|l|l|l|l|l}
\shline
\multirow{2}{*}{Method} & \multirow{2}{*}{Pretraining Dataset} &\multicolumn{2}{c|}{Classification [AUC (\%)]} & \multicolumn{2}{c}{Segmentation [Dice (\%)]}  \\  
\cline{3-4}
\cline{5-6}
     && ChestX-ray14 & CheXpert & SIIM-ACR & Montgomery \\
\shline
    Random & - & 80.31$\pm$0.10 &	86.62$\pm$0.15&	67.54$\pm$0.60	&97.55$\pm$0.36\\
    \hline
    Supervised & ImageNet & \textbf{81.70$\pm$0.15}&	87.17$\pm$0.22&	67.93$\pm$1.45&	\underline{98.19$\pm$0.13}\\
    Supervised & ChestX-ray14 & -& 87.40$\pm$0.26&	68.92$\pm$0.98&	98.16$\pm$0.05 \\
    \hline
    DiRA$_{\text{MoCo-v2}}$& ChestX-ray14 & \underline{81.12$\pm$0.17} &	\textbf{87.59$\pm$0.28}~$\orange{\uparrow}$ ~$\blue{\uparrow}$&	\underline{69.24$\pm$0.41}~$\orange{\uparrow}$ ~$\blue{*}$&	\textbf{98.24$\pm$0.09} ~$\orange{*}$ ~$\blue{\uparrow}$\\
    DiRA$_{\text{Barlow Twins}}$ & ChestX-ray14& 80.88$\pm$0.30	&\underline{87.50$\pm$0.27}~$\orange{\uparrow}$ ~$\blue{*}$&	\textbf{69.87$\pm$0.68}~$\orange{\uparrow}$ ~$\blue{\uparrow}$&	98.16$\pm$0.06 ~$\orange{*}$ ~$\blue{*}$\\
   DiRA$_{\text{SimSiam}}$ & ChestX-ray14& 80.44$\pm$0.29&	86.04$\pm$0.43&	68.76$\pm$0.69~$\orange{*}$ ~$\blue{*}$&	98.17$\pm$0.11 ~$\orange{*}$ ~$\blue{*}$\\   \shline
\end{tabular}
}
\caption{\textbf{Comparison with fully-supervised transfer learning:} DiRA models outperform fully-supervised pretrained models on ImageNet and ChestX-ray14 in three downstream tasks. The best methods are bolded while the second best are underlined. $\orange{\uparrow}$ and $\blue{\uparrow}$ present the statistically significant ($p < 0.05$) improvement compared with supervised ImageNet and ChestX-ray14 baselines, respectively, while  ~$\orange{*}$ and ~$\blue{*}$ presents the statistically equivalent performances accordingly. For supervised ChestX-ray14 model, transfer learning to ChestX-ray14 is not applicable since pretraining and downstream tasks are the same, denoted by “-”.}
\label{tab:2d_sota}
\end{table*} 

\medskip
\noindent\textbf{Results:} As shown in \cref{tab:annotation}, our DiRA pretrained models outperform their counterparts' original methods in all subsets, 1\%, 25\%, and 50\%, across ChestX-ray14, CheXpert, and Montgomery. In particular, the average of improvement for MoCo-v2 and SimSiam across all three downstream tasks in each underlying subset garnering: (1) 5.6 \% and 7\% when using 1\%, (2) 2.9 \% and 1.3\% when using 25\%, and (3) 2.2 \% and 1\% when using 50\%.
As seen in 1\%, DiRA outperforms its counterparts MoCo-v2 and SimSiam by a large margin, demonstrating our framework's potential for combating overfitting in extreme low data regimes. Although the Barlow Twins is more resistant to low data regimes than the previous two approaches, DiRA still improves its performance by 0.5\%, 0.5\%, and 0.6\% on average across all three datasets when using 1\%, 25\%, and 50\% of labeled data, respectively.
In summary, our results in the low-data regimes demonstrate our framework's superiority for providing more robust and transferable representations that can be harnessed for downstream tasks with limited amounts of data, thereby reducing annotation costs.


\subsection{DiRA improves weakly-supervised localization}
\label{sec:weakly_supervised}
\noindent\textbf{Experimental setup:} We investigate our DiRA framework in a weakly supervised setting, comparing its applicability for localizing chest pathology to underlying discriminative methods. Given this goal, we follow~\cite{wang2017chestx} and use the ChestX-ray14 dataset, which contains bounding box annotations for approximately 1,000 images. For training, we initialize models with our DiRA pretrained models, and train downstream models using only image-level disease labels. Following~\cite{wang2017chestx,Selvaraju2017GradCAM}, bounding boxes are only used as  ground truth to evaluate  disease localization accuracy in the testing phase. To generate  heatmaps, we leverage Grad-CAM~\cite{Selvaraju2017GradCAM}. Heatmaps indicate the spatial location of a particular thoracic disease.

\noindent\textbf{Results:}  As seen in \cref{fig:CAM}, our framework learns more fine-grained representations, enabling it to localize diseases more accurately.  In particular, heatmaps generated by MoCo-v2, Barlow Twins, and SimSiam models are highly variable, whereas DiRA models consistently achieve more robust and accurate localization results over each corresponding original method. Through the production of more interpretable activation maps, our DiRA framework demonstrates possible clinical potential for post-hoc interpretation by radiologists. Quantitative disease localization results are provided in the Appendix.


\subsection{DiRA outperforms fully-supervised baselines}
\noindent\textbf{Experimental setup:} Following the recent transfer learning benchmark in medical imaging~\cite{Taher2021Systematic}, we compare the transferability of DiRA models, pretrained solely on unlabeled images from ChestX-ray14, with two fully-supervised representation learning approaches: (1) supervised ImageNet model, the most common transfer learning pipeline in medical imaging and (2) supervised model pretrained on ChestX-ray14, the upper-bound in-domain transfer learning baseline. The supervised baselines benefit from the same encoder as DiRA, namely ResNet-50. We fine-tune all pretrained models for 4 distinct medical applications ranging from target tasks on the source dataset to the tasks with comparatively significant domain-shifts in terms of data distribution and disease/object of interest. 

\smallskip
\noindent\textbf{Results:} As shown in~\cref{tab:2d_sota}, DiRA models achieves significantly better or on-par performance compared with both supervised ImageNet and ChestX-ray14 models across four downstream tasks. In particular, DiRA$_{\text{MoCo-v2}}$ and DiRA$_{\text{Barlow Twins}}$, outperforms both supervised baselines in CheXpert, SIIM-ACR,  and Montgomery, respectively. Moreover, DiRA$_{\text{SimSiam}}$ outperforms the supervised ImageNet and the ChestX-ray14 pretrained models in SIIM-ACR and Montgomery, respectively. These results indicate that our framework, with zero annotated data, is capable of providing more generic features for different medical tasks. 

\begin{table}[]
\centering
\scalebox{0.84}{
\begin{tabular}{l|l|l|l}
\shline
\multirow{2}{*}{Dataset} & \multicolumn{3}{c}{Method} \\
       \cline{2-4} &  Random  & TransVW~\cite{haghighi2021transferable} &
       DiRA$_{\text{TransVW}}$ \\ \shline
       LUNA & 94.25$\pm$5.07 &98.46$\pm$0.30&\bf{98.87$\pm$0.61} \gbf{$\uparrow$ 0.41}\\ \hline
        LIDC-IDRI & 74.05$\pm$1.97&77.33$\pm$0.52 &\textbf{77.51$\pm$1.36} \gbf{$\uparrow$ 0.18}\\ \hline
       LiTS & 79.76$\pm$5.42&86.53$\pm$1.30 & \textbf{86.85$\pm$0.81} \gbf{$\uparrow$ 0.32}\\ \hline
       BraTS &59.87$\pm$4.04 &68.82$\pm$0.38 &\textbf{69.57$\pm$1.13} \gbf{$\uparrow$ 0.75}\\ \hline
        PE-CAD& 80.36$\pm$3.58& \underline{87.07$\pm$2.83} & \underline{86.91$\pm$3.27}\\ \shline
\end{tabular}
}
\caption{\textbf{Comparison with  restorative self-supervised method:} We apply our DiRA to the TransVW as the SOTA restorative self-supervised method. DiRA enhances TransVW by conserving more fine-grained details, resulting in performance boosts in four 3D downstream tasks.}
\label{tab:sota_3D}
\end{table} 

\begin{table*}[]
\centering
\scalebox{0.7}{
\begin{tabular}{l|l|l l l |l|l|l|l}
\shline
\multirow{2}{*}{Base} &
\multirow{2}{*}{Pretraining dataset} &\multirow{2}{*}{$\mathcal{L}_{dis}$} & \multirow{2}{*}{$\mathcal{L}_{res}$} &
\multirow{2}{*}{$\mathcal{L}_{adv}$}&\multicolumn{2}{c|}{Classification [AUC (\%)]} & \multicolumn{2}{c}{Segmentation [Dice (\%)]} \\
\cline{6-7}
\cline{8-9}
       &  & &  &  & ChestX-ray14 & CheXpert & SIIM-ACR &
Montgomery \\

\shline
\multirow{3}{*}{MoCo-v2}& \multirow{3}{*}{ChestX-ray14}&\checkmark & $\times$ & $\times$&80.36$\pm$0.26 & 86.42$\pm$0.42 &67.89$\pm$1.14&98.03$\pm$0.22\\
& &\checkmark &  \checkmark & $\times$ & 80.72$\pm$0.29~$\green{\pmb{\uparrow}}$  &86.86$\pm$0.37~$\green{\pmb{\uparrow}}$& 68.16$\pm$1.07~$\green{\pmb{\uparrow}}$&98.19$\pm$0.08~$\green{\pmb{\uparrow}}$\\
& &\checkmark &  \checkmark & \checkmark & 81.12$\pm$0.17~$\green{\pmb{\uparrow}}$&87.59$\pm$0.28~$\green{\pmb{\uparrow}}$ &69.24$\pm$0.41~$\green{\pmb{\uparrow}}$&98.24$\pm$0.09~$\green{\pmb{\uparrow}}$\\
\hline
\multirow{3}{*}{Barlow Twins}& \multirow{3}{*}{ChestX-ray14}&\checkmark & $\times$ & $\times$ & 80.45$\pm$0.29&86.90$\pm$0.62&69.71$\pm$0.34&98.13$\pm$0.13\\
& &\checkmark &  \checkmark & $\times$& 80.86$\pm$0.16~$\green{\pmb{\uparrow}}$&87.44$\pm$0.33~$\green{\pmb{\uparrow}}$&69.83$\pm$0.29~$\green{\pmb{\uparrow}}$&98.15$\pm$0.14~$\green{\pmb{\uparrow}}$ \\
& &\checkmark &  \checkmark & \checkmark &80.88$\pm$0.30~$\green{\pmb{\uparrow}}$&87.50$\pm$0.27~$\green{\pmb{\uparrow}}$&69.87$\pm$0.68~$\green{\pmb{\uparrow}}$&98.16$\pm$0.06~$\green{\pmb{\uparrow}}$ \\
\hline
\multirow{3}{*}{SimSiam}& \multirow{3}{*}{ChestX-ray14}& \checkmark & $\times$ & $\times$ &79.62$\pm$0.34 &83.82$\pm$0.94&67.58$\pm$1.89 &97.72$\pm$0.27\\
& &\checkmark &  \checkmark & $\times$ &79.41$\pm$0.42~$\gray{\pmb{\downarrow}}$&84.45$\pm$0.46~$\green{\pmb{\uparrow}}$ &68.35$\pm$1.16~$\green{\pmb{\uparrow}}$&98.02$\pm$0.21~$\green{\pmb{\uparrow}}$\\
& & \checkmark &  \checkmark & \checkmark &80.44$\pm$0.29~$\green{\pmb{\uparrow}}$&86.04$\pm$0.43~$\green{\pmb{\uparrow}}$&68.76$\pm$0.69~$\green{\pmb{\uparrow}}$&98.17$\pm$0.11~$\green{\pmb{\uparrow}}$\\ \shline
\end{tabular}
}
\caption{\textbf{Ablation study on different components of DiRA:} We study the impact of each component of DiRA, including discrimination, restoration, and adversary, in four downstream tasks. Adding restorative learning ($\mathcal{L}_{res}$) to discriminative learning leads to consistent performance improvements. Furthermore, equipping models with adversarial learning ($\mathcal{L}_{adv}$) yields performance boosts across all tasks. }
\label{tab:abblation_2D}
\end{table*}

\subsection{DiRA  sets a new state-of-the-art for self-supervised learning in 3D medical imaging}
\noindent\textbf{Experimental setup:} We further investigate the effectiveness of our framework for enhancing restorative representation learning by applying DiRA to TransVW~\cite{haghighi2021transferable}, the state-of-the-art SSL approach for 3D medical imaging. We select TransVW as representative of restorative self-supervised methods because it shows superior performance over discriminative~\cite{Zhuang2019Self,Taleb2020Self}, restorative only~\cite{ZHOU2021Models,chen2019self}, and restorative and adversarial~\cite{Tao2020Revisiting} methods . Following the common evaluation pipeline~\cite{haghighi2021transferable}, we evaluate our learned representations by transfer learning to five common and challenging 3D downstream tasks, including classification (LUNA and PE-CAD) and segmentation (LIDC, LiTS, and BraTS).

\smallskip
\noindent\textbf{Results:} 
As shown in \cref{tab:sota_3D}, DiRA framework consistently enhances TransVW  across all downstream tasks. In particular, DiRA improves TransVW in LUNA, LIDC-IDRI, LiTS, and BraTS, and offers equivalent performance in PE-CAD.  These results imply that by utilizing three learning components in tandem, image-based self-supervision approaches capture a more diverse visual representation that generalizes better to different downstream tasks.

\section{Ablation study}
\label{sec:ablation}
\noindent\textbf{Experimental setup:} We conduct a thorough ablation study to show how each component contributes to DiRA. To do so, we only vary the loss function of DiRA. For each underlying self-supervised method, i.e. MoCo-v2, Barlow Twins, and SimSiam (referred to as the base), we start with the discrimination component and incrementally add  restorative and the adversarial learning. When all three components are unified, they represent the completed DiRA models. All models are pretrained on the ChestX-ray14 dataset and fine-tuned for four downstream tasks, including ChestX-ray14, CheXpert, SIIM-ACR, and Montgomery.

\noindent\textbf{Results:} We draw the following observations from the results in \cref{tab:abblation_2D}: (1) Expanding discriminative self-supervised methods by adding a restoration task consistently enhances the original methods. In particular, incorporating $\mathcal{L}_{res}$ into training objectives of MoCo-v2, Barlow Twins, and SimSiam outperforms the corresponding original methods, with the exception of SimSiam in ChestX-ray14, which shows slight performance degradation. Note that this gap later compensates after adding $\mathcal{L}_{adv}$, which signifies collaborative learning among restorative and adversary components  in our framework. (2) The overall trend showcases the advantage of the adversarial discriminator when added to the restoration component, improving the performance of all methods in four downstream tasks. Our findings indicate that unifying the three components in DiRA models significantly enhances the original self-supervised methods by retaining more fine-grained information from images. 

\section{Conclusion and discussion}
We propose DiRA, the \textit{first} SSL framework that unites discriminative, restorative, and adversarial learning in a unified manner. The key contribution of our DiRA arises from the insights that we have gained into the \textit{synergy} of these three SSL approaches for collaborative learning. Given DiRA’s generalizability, we envisage it will take a fundamental step towards developing universal representations for medical imaging. Our DiRA achieves remarkable performance gains, though we fixed the restorative learning tasks in all experiments when examining various formulations of discriminative learning. In the future, examining various choices of restoration tasks and searching for optimal collaborative learning strategies may lead to even stronger representations for medical imaging. In this paper, we have focused on medical imaging, but we envision that DiRA can also offer outstanding performance for vision tasks that demand fine-grained details.

\noindent\textbf{Acknowledgments:}
With the help of Zongwei Zhou, Zuwei Guo started implementing the earlier ideas behind ``United \& Unified'', which has branched out  into DiRA. We thank them for their feasibility exploration, especially their initial evaluation on TransVW\cite{haghighi2021transferable} and various training strategies. This research has been supported in part by ASU and Mayo Clinic through a Seed Grant and an Innovation Grant and in part by the NIH under Award Number R01HL128785. The content is solely the responsibility of the authors and does not necessarily represent the official views of the NIH. This work  utilized the GPUs provided in part by the ASU Research Computing and in part by the Extreme Science and Engineering Discovery Environment (XSEDE) funded by the National Science Foundation (NSF) under grant number ACI-1548562.  Paper content is covered by patents pending.


\begin{thebibliography}{10}\itemsep=-1pt
\bibitem{PNEchallenge}
Siim-acr pneumothorax segmentation, 2019.

\bibitem{armato2011lung}
Samuel~G Armato~III, Geoffrey McLennan, Luc Bidaut, Michael~F McNitt-Gray,
  Charles~R Meyer, Anthony~P Reeves, Binsheng Zhao, Denise~R Aberle, Claudia~I
  Henschke, Eric~A Hoffman, et~al.
\newblock The lung image database consortium (lidc) and image database resource
  initiative (idri): a completed reference database of lung nodules on ct
  scans.
\newblock {\em Medical physics}, 38(2):915--931, 2011.

\bibitem{azizi2021big}
Shekoofeh Azizi, Basil Mustafa, Fiona Ryan, Zachary Beaver, Jan Freyberg,
  Jonathan Deaton, Aaron Loh, Alan Karthikesalingam, Simon Kornblith, Ting
  Chen, Vivek Natarajan, and Mohammad Norouzi.
\newblock Big self-supervised models advance medical image classification.
\newblock {\em arXiv:2101.05224}, 2021.

\bibitem{bakas2018identifying}
Spyridon Bakas, Mauricio Reyes, Andras Jakab, Stefan Bauer, Markus Rempfler,
  Alessandro Crimi, Russell~Takeshi Shinohara, Christoph Berger, Sung~Min Ha,
  Martin Rozycki, et~al.
\newblock Identifying the best machine learning algorithms for brain tumor
  segmentation, progression assessment, and overall survival prediction in the
  brats challenge.
\newblock {\em arXiv:1811.02629}, 2018.

\bibitem{bilic2019liver}
Patrick Bilic, Patrick~Ferdinand Christ, Eugene Vorontsov, Grzegorz Chlebus,
  Hao Chen, Qi Dou, Chi-Wing Fu, Xiao Han, Pheng-Ann Heng, J{\"u}rgen Hesser,
  et~al.
\newblock The liver tumor segmentation benchmark (lits).
\newblock {\em arXiv:1901.04056}, 2019.

\bibitem{CaoAuto2020}
Bing Cao, Han Zhang, Nannan Wang, Xinbo Gao, and Dinggang Shen.
\newblock Auto-gan: Self-supervised collaborative learning for medical image
  synthesis.
\newblock {\em Proceedings of the AAAI Conference on Artificial Intelligence},
  34(07):10486--10493, Apr. 2020.

\bibitem{caron2018deep}
Mathilde Caron, Piotr Bojanowski, Armand Joulin, and Matthijs Douze.
\newblock Deep clustering for unsupervised learning of visual features.
\newblock In {\em Proceedings of the European Conference on Computer Vision},
  pages 132--149, 2018.

\bibitem{caron2021unsupervised}
Mathilde Caron, Ishan Misra, Julien Mairal, Priya Goyal, Piotr Bojanowski, and
  Armand Joulin.
\newblock Unsupervised learning of visual features by contrasting cluster
  assignments.
\newblock {\em arXiv:2006.09882}, 2021.

\bibitem{Chaitanya2020Contrastive}
Krishna Chaitanya, Ertunc Erdil, Neerav Karani, and Ender Konukoglu.
\newblock Contrastive learning of global and local features for medical image
  segmentation with limited annotations.
\newblock In {\em Advances in Neural Information Processing Systems},
  volume~33, pages 12546--12558. Curran Associates, Inc., 2020.

\bibitem{chen2019self}
Liang Chen, Paul Bentley, Kensaku Mori, Kazunari Misawa, Michitaka Fujiwara,
  and Daniel Rueckert.
\newblock Self-supervised learning for medical image analysis using image
  context restoration.
\newblock {\em Medical image analysis}, 58:101539, 2019.

\bibitem{Chen2020Generative}
Mark Chen, Alec Radford, Rewon Child, Jeffrey Wu, Heewoo Jun, David Luan, and
  Ilya Sutskever.
\newblock Generative pretraining from pixels.
\newblock In Hal~Daumé III and Aarti Singh, editors, {\em Proceedings of the
  37th International Conference on Machine Learning}, volume 119 of {\em
  Proceedings of Machine Learning Research}, pages 1691--1703. PMLR, 13--18 Jul
  2020.

\bibitem{Chen2020Simple}
Ting Chen, Simon Kornblith, Mohammad Norouzi, and Geoffrey Hinton.
\newblock A simple framework for contrastive learning of visual
  representations.
\newblock In Hal~Daumé III and Aarti Singh, editors, {\em Proceedings of the
  37th International Conference on Machine Learning}, volume 119 of {\em
  Proceedings of Machine Learning Research}, pages 1597--1607. PMLR, 13--18 Jul
  2020.

\bibitem{chen2020big}
Ting Chen, Simon Kornblith, Kevin Swersky, Mohammad Norouzi, and Geoffrey
  Hinton.
\newblock Big self-supervised models are strong semi-supervised learners, 2020.

\bibitem{chen2020improved}
Xinlei Chen, Haoqi Fan, Ross Girshick, and Kaiming He.
\newblock Improved baselines with momentum contrastive learning, 2020.

\bibitem{Chen2021Exploring}
Xinlei Chen and Kaiming He.
\newblock Exploring simple siamese representation learning.
\newblock In {\em Proceedings of the IEEE/CVF Conference on Computer Vision and
  Pattern Recognition (CVPR)}, pages 15750--15758, June 2021.

\bibitem{devries2017improved}
Terrance DeVries and Graham~W. Taylor.
\newblock Improved regularization of convolutional neural networks with cutout,
  2017.

\bibitem{doersch2015Unsupervised}
Carl Doersch, Abhinav Gupta, and Alexei~A Efros.
\newblock Unsupervised visual representation learning by context prediction.
\newblock In {\em Proceedings of the IEEE International Conference on Computer
  Vision}, pages 1422--1430, 2015.

\bibitem{Donahue2019Large}
Jeff Donahue and Karen Simonyan.
\newblock Large scale adversarial representation learning.
\newblock In H. Wallach, H. Larochelle, A. Beygelzimer, F. d\textquotesingle
  Alch\'{e}-Buc, E. Fox, and R. Garnett, editors, {\em Advances in Neural
  Information Processing Systems}, volume~32. Curran Associates, Inc., 2019.

\bibitem{Dumoulin2017Adversarially}
Vincent Dumoulin, Ishmael Belghazi, Ben Poole, Alex Lamb, Mart{\'{\i}}n
  Arjovsky, Olivier Mastropietro, and Aaron~C. Courville.
\newblock Adversarially learned inference.
\newblock In {\em 5th International Conference on Learning Representations,
  {ICLR} 2017, Toulon, France, April 24-26, 2017, Conference Track
  Proceedings}. OpenReview.net, 2017.

\bibitem{Ellis2021Trialing}
David~G. Ellis and Michele~R. Aizenberg.
\newblock Trialing u-net training modifications for segmenting gliomas using
  open source deep learning framework.
\newblock In Alessandro Crimi and Spyridon Bakas, editors, {\em Brainlesion:
  Glioma, Multiple Sclerosis, Stroke and Traumatic Brain Injuries}, pages
  40--49, Cham, 2021. Springer International Publishing.

\bibitem{Ermolov2021Whitening}
Aleksandr Ermolov, Aliaksandr Siarohin, Enver Sangineto, and Nicu Sebe.
\newblock Whitening for self-supervised representation learning.
\newblock In Marina Meila and Tong Zhang, editors, {\em Proceedings of the 38th
  International Conference on Machine Learning}, volume 139 of {\em Proceedings
  of Machine Learning Research}, pages 3015--3024. PMLR, 18--24 Jul 2021.

\bibitem{gidaris2018unsupervised}
Spyros Gidaris, Praveer Singh, and Nikos Komodakis.
\newblock Unsupervised representation learning by predicting image rotations.
\newblock {\em arXiv:1803.07728}, 2018.

\bibitem{NIPS2014Goodfellow}
Ian Goodfellow, Jean Pouget-Abadie, Mehdi Mirza, Bing Xu, David Warde-Farley,
  Sherjil Ozair, Aaron Courville, and Yoshua Bengio.
\newblock Generative adversarial nets.
\newblock In Z. Ghahramani, M. Welling, C. Cortes, N. Lawrence, and K.~Q.
  Weinberger, editors, {\em Advances in Neural Information Processing Systems},
  volume~27. Curran Associates, Inc., 2014.

\bibitem{Grill2020Bootstrap}
Jean-Bastien Grill, Florian Strub, Florent Altch\'{e}, Corentin Tallec, Pierre
  Richemond, Elena Buchatskaya, Carl Doersch, Bernardo Avila~Pires, Zhaohan
  Guo, Mohammad Gheshlaghi~Azar, Bilal Piot, koray kavukcuoglu, Remi Munos, and
  Michal Valko.
\newblock Bootstrap your own latent - a new approach to self-supervised
  learning.
\newblock In H. Larochelle, M. Ranzato, R. Hadsell, M.~F. Balcan, and H. Lin,
  editors, {\em Advances in Neural Information Processing Systems}, volume~33,
  pages 21271--21284. Curran Associates, Inc., 2020.

\bibitem{haghighi2020learning}
Fatemeh Haghighi, Mohammad~Reza Hosseinzadeh~Taher, Zongwei Zhou, Michael~B.
  Gotway, and Jianming Liang.
\newblock Learning semantics-enriched representation via self-discovery,
  self-classification, and self-restoration.
\newblock In {\em Medical Image Computing and Computer Assisted Intervention --
  MICCAI 2020}, pages 137--147, Cham, 2020. Springer International Publishing.

\bibitem{haghighi2021transferable}
Fatemeh Haghighi, Mohammad Reza~Hosseinzadeh Taher, Zongwei Zhou, Michael~B.
  Gotway, and Jianming Liang.
\newblock Transferable visual words: Exploiting the semantics of anatomical
  patterns for self-supervised learning.
\newblock {\em IEEE Transactions on Medical Imaging}, 40(10):2857--2868, 2021.

\bibitem{He2020Momentum}
Kaiming He, Haoqi Fan, Yuxin Wu, Saining Xie, and Ross Girshick.
\newblock Momentum contrast for unsupervised visual representation learning.
\newblock In {\em Proceedings of the IEEE/CVF Conference on Computer Vision and
  Pattern Recognition (CVPR)}, June 2020.

\bibitem{he2016deep}
Kaiming He, Xiangyu Zhang, Shaoqing Ren, and Jian Sun.
\newblock Deep residual learning for image recognition.
\newblock In {\em Proceedings of the IEEE Conference on Computer Vision and
  Pattern Recognition}, pages 770--778, 2016.

\bibitem{Taher2021Systematic}
Mohammad~Reza Hosseinzadeh~Taher, Fatemeh Haghighi, Ruibin Feng, Michael~B.
  Gotway, and Jianming Liang.
\newblock A systematic benchmarking analysis of transfer learning for medical
  image analysis.
\newblock In {\em Domain Adaptation and Representation Transfer, and Affordable
  Healthcare and AI for Resource Diverse Global Health}, pages 3--13, Cham,
  2021. Springer International Publishing.

\bibitem{irvin2019chexpert}
Jeremy Irvin, Pranav Rajpurkar, Michael Ko, Yifan Yu, Silviana Ciurea-Ilcus,
  Chris Chute, Henrik Marklund, Behzad Haghgoo, Robyn Ball, Katie Shpanskaya,
  et~al.
\newblock Chexpert: A large chest radiograph dataset with uncertainty labels
  and expert comparison.
\newblock {\em arXiv:1901.07031}, 2019.

\bibitem{Jaeger2014Tow}
Stefan Jaeger, Sema Candemir, Sameer Antani, Yì-Xiáng~J Wáng, Pu-Xuan Lu,
  and George Thoma.
\newblock Two public chest x-ray datasets for computer-aided screening of
  pulmonary diseases.
\newblock {\em Quantitative imaging in medicine and surgery}, 4(6), 2014.

\bibitem{larsson2017colorproxy}
Gustav Larsson, Michael Maire, and Gregory Shakhnarovich.
\newblock Colorization as a proxy task for visual understanding.
\newblock In {\em CVPR}, 2017.

\bibitem{li2021prototypical}
Junnan Li, Pan Zhou, Caiming Xiong, and Steven C.~H. Hoi.
\newblock Prototypical contrastive learning of unsupervised representations,
  2021.

\bibitem{noroozi2016unsupervised}
Mehdi Noroozi and Paolo Favaro.
\newblock Unsupervised learning of visual representations by solving jigsaw
  puzzles.
\newblock In {\em European Conference on Computer Vision}, pages 69--84.
  Springer, 2016.

\bibitem{Parmar2021Dual}
Gaurav Parmar, Dacheng Li, Kwonjoon Lee, and Zhuowen Tu.
\newblock Dual contradistinctive generative autoencoder.
\newblock In {\em Proceedings of the IEEE/CVF Conference on Computer Vision and
  Pattern Recognition (CVPR)}, pages 823--832, June 2021.

\bibitem{pathak2016context}
Deepak Pathak, Philipp Krahenbuhl, Jeff Donahue, Trevor Darrell, and Alexei~A
  Efros.
\newblock Context encoders: Feature learning by inpainting.
\newblock In {\em Proceedings of the IEEE Conference on Computer Vision and
  Pattern Recognition}, pages 2536--2544, 2016.

\bibitem{Ronneberger2015Unet}
Olaf Ronneberger, Philipp Fischer, and Thomas Brox.
\newblock U-net: Convolutional networks for biomedical image segmentation.
\newblock In Nassir Navab, Joachim Hornegger, William~M. Wells, and
  Alejandro~F. Frangi, editors, {\em Medical Image Computing and
  Computer-Assisted Intervention -- MICCAI 2015}, pages 234--241, Cham, 2015.
  Springer International Publishing.

\bibitem{Selvaraju2017GradCAM}
Ramprasaath~R. Selvaraju, Michael Cogswell, Abhishek Das, Ramakrishna Vedantam,
  Devi Parikh, and Dhruv Batra.
\newblock Grad-cam: Visual explanations from deep networks via gradient-based
  localization.
\newblock In {\em Proceedings of the IEEE International Conference on Computer
  Vision (ICCV)}, Oct 2017.

\bibitem{setio2017validation}
Arnaud Arindra~Adiyoso Setio, Alberto Traverso, Thomas De~Bel, Moira~SN Berens,
  Cas van~den Bogaard, Piergiorgio Cerello, Hao Chen, Qi Dou, Maria~Evelina
  Fantacci, Bram Geurts, et~al.
\newblock Validation, comparison, and combination of algorithms for automatic
  detection of pulmonary nodules in computed tomography images: the luna16
  challenge.
\newblock {\em Medical image analysis}, 42:1--13, 2017.

\bibitem{Taher2022CAiD}
Mohammad Reza~Hosseinzadeh Taher, Fatemeh Haghighi, Michael~B. Gotway, and
  Jianming Liang.
\newblock Caid: Context-aware instance discrimination for self-supervised
  learning in medical imaging.
\newblock {\em arXiv:2204.07344}, 2022.

\bibitem{tajbakhsh2015computer}
Nima Tajbakhsh, Michael~B Gotway, and Jianming Liang.
\newblock Computer-aided pulmonary embolism detection using a novel
  vessel-aligned multi-planar image representation and convolutional neural
  networks.
\newblock In {\em International Conference on Medical Image Computing and
  Computer-Assisted Intervention}, pages 62--69. Springer, 2015.

\bibitem{Taleb2020Self}
Aiham Taleb, Winfried Loetzsch, Noel Danz, Julius Severin, Thomas Gaertner,
  Benjamin Bergner, and Christoph Lippert.
\newblock 3d self-supervised methods for medical imaging.
\newblock In H. Larochelle, M. Ranzato, R. Hadsell, M.~F. Balcan, and H. Lin,
  editors, {\em Advances in Neural Information Processing Systems}, volume~33,
  pages 18158--18172. Curran Associates, Inc., 2020.

\bibitem{Tao2020Revisiting}
Xing Tao, Yuexiang Li, Wenhui Zhou, Kai Ma, and Yefeng Zheng.
\newblock Revisiting rubik's cube: Self-supervised learning with volume-wise
  transformation for 3d medical image segmentation.
\newblock In Anne~L. Martel, Purang Abolmaesumi, Danail Stoyanov, Diana Mateus,
  Maria~A. Zuluaga, S.~Kevin Zhou, Daniel Racoceanu, and Leo Joskowicz,
  editors, {\em Medical Image Computing and Computer Assisted Intervention --
  MICCAI 2020}, pages 238--248, Cham, 2020. Springer International Publishing.

\bibitem{tian2020makes}
Yonglong Tian, Chen Sun, Ben Poole, Dilip Krishnan, Cordelia Schmid, and
  Phillip Isola.
\newblock What makes for good views for contrastive learning?, 2020.

\bibitem{Vincent2008Extracting}
Pascal Vincent, Hugo Larochelle, Yoshua Bengio, and Pierre-Antoine Manzagol.
\newblock Extracting and composing robust features with denoising autoencoders.
\newblock In {\em Proceedings of the 25th International Conference on Machine
  Learning}, ICML ’08, page 1096–1103. Association for Computing Machinery,
  2008.

\bibitem{wang2017chestx}
Xiaosong Wang, Yifan Peng, Le Lu, Zhiyong Lu, Mohammadhadi Bagheri, and
  Ronald~M Summers.
\newblock Chestx-ray8: Hospital-scale chest x-ray database and benchmarks on
  weakly-supervised classification and localization of common thorax diseases.
\newblock In {\em Proceedings of the IEEE Conference on Computer Vision and
  Pattern Recognition}, pages 2097--2106, 2017.

\bibitem{Wang2021Dense}
Xinlong Wang, Rufeng Zhang, Chunhua Shen, Tao Kong, and Lei Li.
\newblock Dense contrastive learning for self-supervised visual pre-training.
\newblock In {\em Proceedings of the IEEE/CVF Conference on Computer Vision and
  Pattern Recognition (CVPR)}, pages 3024--3033, June 2021.

\bibitem{Wu2018Unsupervised}
Zhirong Wu, Yuanjun Xiong, Stella~X. Yu, and Dahua Lin.
\newblock Unsupervised feature learning via non-parametric instance
  discrimination.
\newblock In {\em Proceedings of the IEEE Conference on Computer Vision and
  Pattern Recognition (CVPR)}, June 2018.

\bibitem{Wu2018insdis}
Zhirong Wu, Yuanjun Xiong, Stella~X Yu, and Dahua Lin.
\newblock Unsupervised feature learning via non-parametric instance
  discrimination.
\newblock In {\em Proceedings of the IEEE conference on computer vision and
  pattern recognition}, pages 3733--3742, 2018.

\bibitem{Xie2021DetCo}
Enze Xie, Jian Ding, Wenhai Wang, Xiaohang Zhan, Hang Xu, Peize Sun, Zhenguo
  Li, and Ping Luo.
\newblock Detco: Unsupervised contrastive learning for object detection.
\newblock In {\em Proceedings of the IEEE/CVF International Conference on
  Computer Vision (ICCV)}, pages 8392--8401, October 2021.

\bibitem{Xie2021Propagate}
Zhenda Xie, Yutong Lin, Zheng Zhang, Yue Cao, Stephen Lin, and Han Hu.
\newblock Propagate yourself: Exploring pixel-level consistency for
  unsupervised visual representation learning.
\newblock In {\em Proceedings of the IEEE/CVF Conference on Computer Vision and
  Pattern Recognition (CVPR)}, pages 16684--16693, June 2021.

\bibitem{Ye2019Unsupervised}
Mang Ye, Xu Zhang, Pong~C. Yuen, and Shih-Fu Chang.
\newblock Unsupervised embedding learning via invariant and spreading instance
  feature.
\newblock In {\em Proceedings of the IEEE/CVF Conference on Computer Vision and
  Pattern Recognition (CVPR)}, June 2019.

\bibitem{zbontar2021barlow}
Jure Zbontar, Li Jing, Ishan Misra, Yann LeCun, and St{\'e}phane Deny.
\newblock Barlow twins: Self-supervised learning via redundancy reduction.
\newblock {\em arXiv:2103.03230}, 2021.

\bibitem{Zhan2020Online}
Xiaohang Zhan, Jiahao Xie, Ziwei Liu, Yew-Soon Ong, and Chen~Change Loy.
\newblock Online deep clustering for unsupervised representation learning.
\newblock In {\em Proceedings of IEEE/CVF Conference on Computer Vision and
  Pattern Recognition (CVPR)}, 2020.

\bibitem{Zhou2021Preservational}
Hong-Yu Zhou, Chixiang Lu, Sibei Yang, Xiaoguang Han, and Yizhou Yu.
\newblock Preservational learning improves self-supervised medical image models
  by reconstructing diverse contexts.
\newblock In {\em Proceedings of the IEEE/CVF International Conference on
  Computer Vision (ICCV)}, pages 3499--3509, October 2021.

\bibitem{Zhou2020Comparing}
Hong-Yu Zhou, Shuang Yu, Cheng Bian, Yifan Hu, Kai Ma, and Yefeng Zheng.
\newblock Comparing to learn: Surpassing imagenet pretraining on radiographs by
  comparing image representations.
\newblock In Anne~L. Martel, Purang Abolmaesumi, Danail Stoyanov, Diana Mateus,
  Maria~A. Zuluaga, S.~Kevin Zhou, Daniel Racoceanu, and Leo Joskowicz,
  editors, {\em Medical Image Computing and Computer Assisted Intervention --
  MICCAI 2020}, pages 398--407, Cham, 2020. Springer International Publishing.

\bibitem{ZHOU2021Models}
Zongwei Zhou, Vatsal Sodha, Jiaxuan Pang, Michael~B. Gotway, and Jianming
  Liang.
\newblock Models genesis.
\newblock {\em Medical Image Analysis}, 67:101840, 2021.

\bibitem{Zhuang2019Self}
Xinrui Zhuang, Yuexiang Li, Yifan Hu, Kai Ma, Yujiu Yang, and Yefeng Zheng.
\newblock Self-supervised feature learning for 3d medical images by playing a
  rubik's cube.
\newblock In Dinggang Shen, Tianming Liu, Terry~M. Peters, Lawrence~H. Staib,
  Caroline Essert, Sean Zhou, Pew-Thian Yap, and Ali Khan, editors, {\em
  Medical Image Computing and Computer Assisted Intervention -- MICCAI 2019},
  pages 420--428, Cham, 2019. Springer International Publishing.
  
\bibitem{Choe2020Evaluating}
Junsuk Choe, Seong~Joon Oh, Seungho Lee, Sanghyuk Chun, Zeynep Akata, and
  Hyunjung Shim.
\newblock Evaluating weakly supervised object localization methods right.
\newblock In {\em Proceedings of the IEEE/CVF Conference on Computer Vision and
  Pattern Recognition (CVPR)}, June 2020.

\bibitem{oord2019representation}
Aaron van~den Oord, Yazhe Li, and Oriol Vinyals.
\newblock Representation learning with contrastive predictive coding, 2019.

\bibitem{Chen2021Joint}
Hao Chen, Yaohui Wang, Benoit Lagadec, Antitza Dantcheva, and Francois Bremond.
\newblock Joint generative and contrastive learning for unsupervised person
  re-identification.
\newblock In {\em Proceedings of the IEEE/CVF Conference on Computer Vision and
  Pattern Recognition (CVPR)}, pages 2004--2013, June 2021.
  
\end{thebibliography}

\clearpage

\appendix
\section*{Appendix}
\begin{table*}[]
\centering
\scalebox{1}{
\begin{tabular}{l|l | l | l |l |l| l}
Method & $\delta=10\%$&$\delta=20\%$&$\delta=30\%$&$\delta=40\%$&$\delta=50\%$&$\delta=60\%$\\ 
\shline
MoCo-v2~\cite{chen2020improved}&  54.89 & 39.43 & 24.81& 14.59& 7.58 & 2.68\\
DiRA$_{\text{MoCo-v2}}$ & \bf{58.13} \gbf{$\uparrow$ 3.2} & \bf{42.74} \gbf{$\uparrow$ 3.3} &\bf{27.52} \gbf{$\uparrow$ 2.7}& \bf{16.25} \gbf{$\uparrow$ 1.7}& \bf{9.30} \gbf{$\uparrow$ 1.7}& \bf{4.35} \gbf{$\uparrow$ 1.7}\\\hline

Barlow Twins~\cite{zbontar2021barlow} &50.54&38.01&26.36&16.93&9.31&4.69 \\
DiRA$_{\text{BarlowTwins}}$ & \bf{58.98} \gbf{$\uparrow$ 8.4}&\bf{45.26} \gbf{$\uparrow$ 7.2}& \bf{32.71} \gbf{$\uparrow$ 6.3} &
\bf{21.71} \gbf{$\uparrow$ 4.8}	&	\bf{13.62} \gbf{$\uparrow$ 4.3}&	\bf{6.26} \gbf{$\uparrow$ 1.6}\\ \hline
SimSiam~\cite{Chen2021Exploring} & 30.24 & 19.80&11.46&5.62&2.30&0.79\\
DiRA$_{\text{SimSiam}}$ & \bf{51.07} \gbf{$\uparrow$ 20.8} & \bf{34.24} \gbf{$\uparrow$ 14.4}& \bf{20.64} \gbf{$\uparrow$ 9.2}&\bf{11.32} \gbf{$\uparrow$ 5.7}	&\bf{6.46} \gbf{$\uparrow$ 4.2}&\bf{2.90} \gbf{$\uparrow$ 2.1}\\
\end{tabular}}
\caption{\textbf{Weakly-supervised pathology localization accuracy under different IoU thresholds ($\delta$):} DiRA models provide stronger representations for pathology localization with only image-level annotations. For each method,  we report the average performance over ten runs. The green arrows show the improvement of DiRA models compared with the underlying discriminative method in each IoU threshold. 
}
\label{tab:weakly_supervised}
\end{table*}

\begin{figure*}[!h]
\centering
    \begin{subfigure}[b]{0.9\textwidth}            
            \includegraphics[width=1\linewidth]{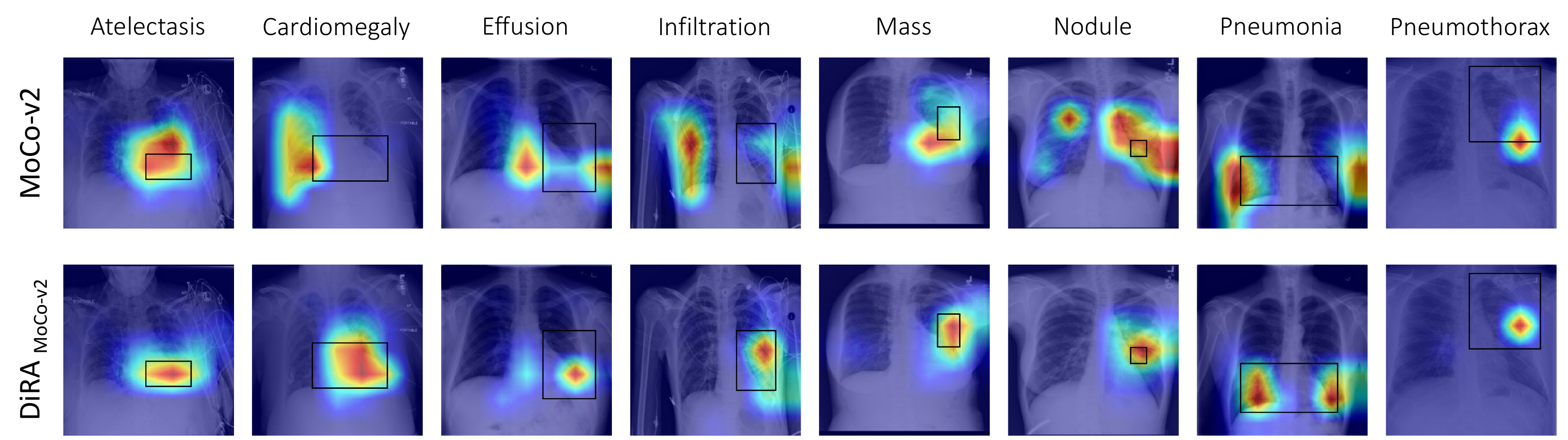}
            \caption{MoCo-v2 vs. DiRA$_{\text{MoCo-v2}}$}
            \label{fig:SRl}
            \vspace{1.2em}
    \end{subfigure}%
    \\
    
    \begin{subfigure}[b]{0.9\textwidth}
            \centering
            \includegraphics[width=1\linewidth]{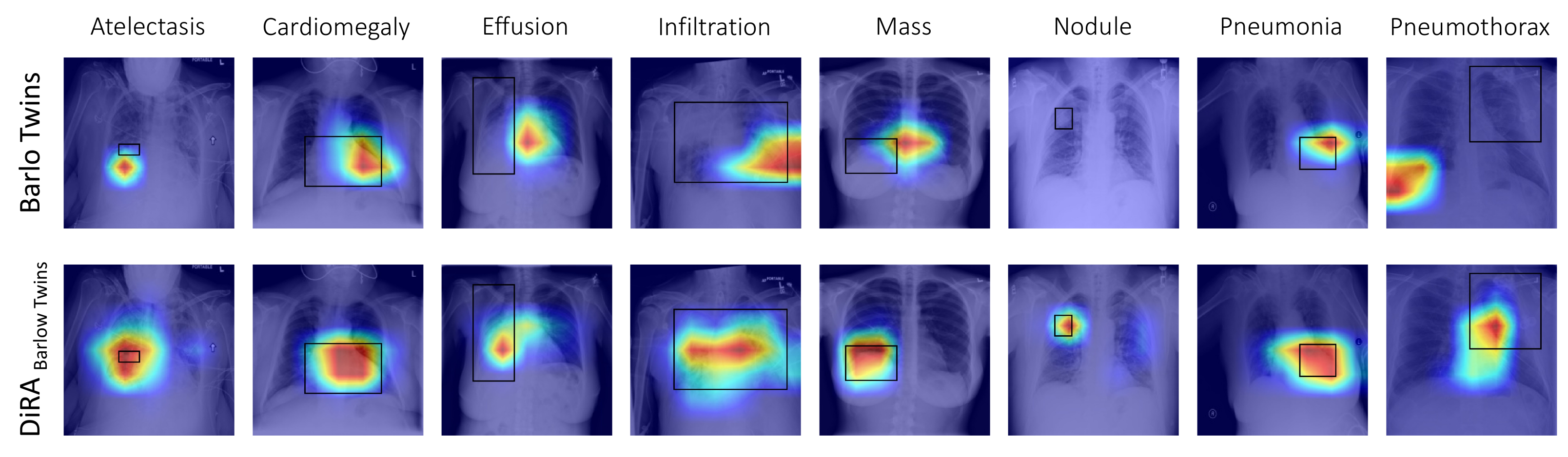}
            \caption{Barlow Twins vs. DiRA$_{\text{Barlow Twins}}$}
            \label{fig:D-Imager}
             \vspace{1.2em}
    \end{subfigure}
        \\
    
    \begin{subfigure}[b]{0.9\textwidth}
            \centering
            \includegraphics[width=1\linewidth]{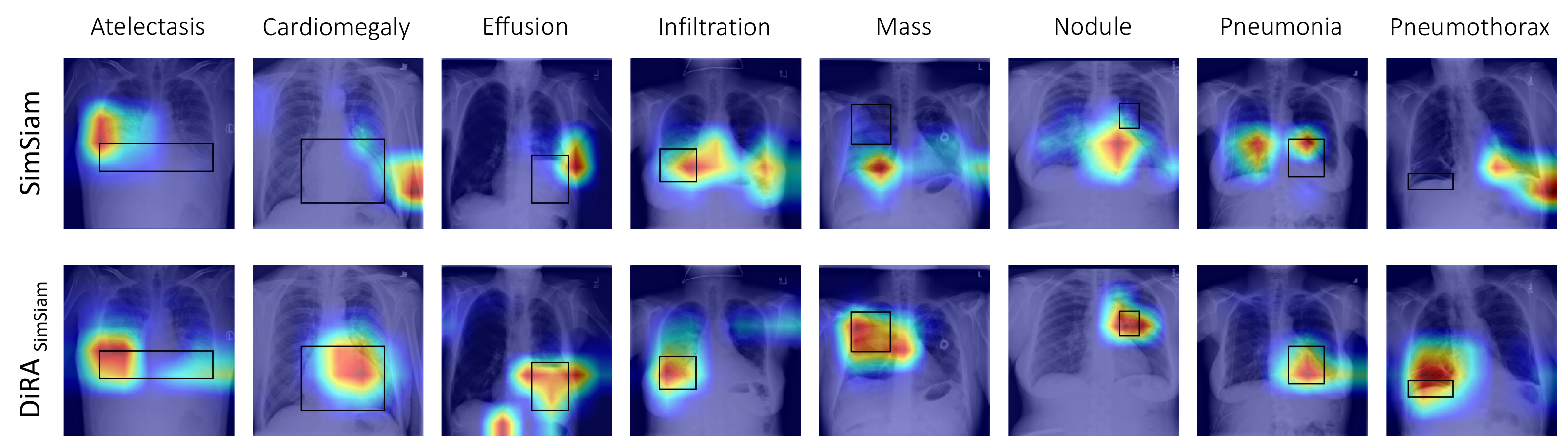}
            \caption{SimSiam vs. DiRA$_{\text{SimSiam}}$}
            \label{fig:D-Imager}
    \end{subfigure}
    
    \caption{\textbf{Visualization of Grad-CAM heatmaps:} We provide the heatmap examples for 8 thorax diseases in each column. The first row in each sub-figure represents the results for the original self-supervised method, while the second row represents the original method when adopted in DiRA framework. The black boxes represents the localization ground truths. }\label{fig:CAMs_appendix}
\end{figure*}

\section{Weakly-supervised localization}
In this section, we provide quantitative and additional qualitative results for weakly-supervised localization, discussed in the \cref{sec:weakly_supervised} of the main paper. Our quantitative results in \cref{tab:weakly_supervised}, together with the qualitative results in \cref{fig:CAM} and \cref{fig:CAMs_appendix}, demonstrate the capability of our framework in learning fine-grained representations that can be used for more accurate pathology localization when just image-level annotations are available.
\subsection{Quantitative results}
\noindent\textbf{Experimental setup:} 
Following the common protocol~\cite{wang2017chestx,Choe2020Evaluating,Selvaraju2017GradCAM}, we  quantitatively evaluate  the  applicability  of our DiRA framework in a weakly supervised setting using ChestX-ray14 dataset. First, we use min-max normalization to normalize each heatmap; then, following~\cite{wang2017chestx}, we binarize the heatmaps by thresholding at \{60, 180\}, and generate bounding boxes around the isolated regions. To evaluate localization accuracy, we compute the intersection over union (IoU) between the generated and ground truth bounding boxes. According to~\cite{wang2017chestx,Choe2020Evaluating}, a localization is correct when the bounding box prediction overlaps with the ground truth box with IoU $\geq \delta$. Following~\cite{wang2017chestx}, we investigate the accuracy of localization under various $\delta$ values, from 10\% to 60\%. We run each method ten times and report the average accuracy across all runs.

\medskip
\noindent\textbf{Result:} \cref{tab:weakly_supervised} shows the pathology localization accuracy of our DiRA  and  underlying  discriminative  models. As seen, in each of the six IoU thresholds, DiRA models significantly outperform the corresponding  discriminative  models. In particular, the average of improvement for MoCo-v2, Barlow Twins, and SimSiam across all IoU thresholds is 2.38\%, 5.4\%, and 9.4\%, respectively. 

\subsection{Qualitative results}
\noindent\textbf{Experimental setup:} 
 During training,  we initialize  models  with  our  DiRA  pre-trained  models,  and  fine-tune downstream models using only image-level disease labels. We use heatmaps to approximate the spatial location of a particular thorax disease. We generate  heatmaps using Grad-CAM~\cite{Selvaraju2017GradCAM}, a technique for highlighting the important regions in the image for predicting the pathology class. 

\medskip
\noindent\textbf{Results:} \cref{fig:CAMs_appendix} presents the visualizations of heatmaps generated by DiRA and the corresponding discriminative models for 8 thorax pathologies in ChestX-ray14 dataset. 
As seen, DiRA models provide more accurate pathology localizations compared to the underlying discriminative methods. These results demonstrate the impact of restorative learning in providing fine-grained features that are useful for disease localization.  

\section{Datasets and tasks} 
We have examined our framework in a diverse suite of 9 downstream tasks, including classification and segmentation in X-ray, CT,  and MRI modalities. In this section, we provide the details of each dataset and the underlying task, as well as the evaluation metric for each task. 

\medskip
\noindent\textbf{ChestX-ray14:} ChestX-ray14 is a large open source dataset of de-identifie chest X-ray images. The dataset includes 112K chest images taken from 30K unique patients. The ground truth consists of a label space of 14 thorax diseases. We use the official patient-wise split released with the dataset, including 86K training images and 25K testing images.  The models are trained to predict 14 pathologies in a multi-label classification setting. The mean AUC score over 14 diseases is used to evaluate the classification performance. In addition to image-level labels, ChestX-ray14 provides bounding box annotations for approximately 1,000 test images. Of this set of images,  bounding box annotations are available for 8 out of 14 thorax diseases. During testing, we use bounding box annotations to assess the accuracy of pathology localization in a weakly-supervised setting. The mean accuracy over 8 diseases is used to evaluate the localization performance.

\medskip
\noindent\textbf{CheXpert:} CheXpert is a hospital-scale publicly available dataset with 224K  chest X-ray images taken from 65K unique patients. We use the official data split released with the dataset, including 224K training and 234 test images. The ground truth for the training set includes 14 thoracic pathologies that were retrieved automatically from radiology reports. The testing set is labeled manually by board-certified radiologists for 5 selected thoracic pathologies--- cardiomegaly, edema, consolidation, atelectasis, and pleural effusion. The models are trained to predict five pathologies in a multi-label classification setting. The mean AUC score over 5 diseases is used to evaluate the classification performance.

\medskip
\noindent\textbf{SIIM-ACR:} This open dataset is provided by the Society for Imaging Informatics in Medicine (SIIM) and American College of Radiology, including 10K chest X-ray images and pixel-wise segmentation mask for Pneumothorax disease.  We randomly divided the dataset into training (80\%) and testing (20\%). The models are trained to segment pneumothorax from chest radiographic images (if present). The segmentation performance was measured by the mean Dice coefficient score. 

\medskip
\noindent\textbf{NIH Montgomery:} This publicly available dataset is provided by the Montgomery County’s Tuberculosis screening program, including 138 chest X-ray images. There are 80 normal cases and 58 cases with Tuberculosis (TB) indications in this dataset. Moreover, ground truth segmentation masks for left and right lungs are provided.  We randomly divided the dataset into a training set (80\%) and a test set (20\%). The models are trained to segment left and right lungs in chest scans. The segmentation performance is evaluated by the mean Dice score.

\medskip
\noindent\textbf{LUNA:} This publicly-available dataset consists of 888 lung CT scans with a slice thickness of less than 2.5mm. The dataset were divided into training  (445 cases),  validation  (178 cases), and  test (265 cases) sets. The dataset provides a set of 5M candidate locations for lung nodule. Each location is labeled as true positive (1) or  false positive (0). The models are trained to classify lung nodule candidates into true positives and false positives in a binary classification setting. We evaluate the classification accuracy by Area Under the Curve (AUC) score.  

\medskip
\noindent\textbf{PE-CAD:} This dataset includes 121 computed tomography pulmonary angiography (CTPA) scans with a total of 326 pulmonary embolism (PE). The dataset provides a set of candidate locations for PE and is divided at the patient-level into training and test sets. Training set contains  434 true positive PE candidates and 3,406 false positive PE candidates. Test set contains 253 true positive PE candidates and 2,162 false positive PE candidates. We pre-processed the 3D scans as suggested in~\cite{ZHOU2021Models}. The 3D models are trained to classify PE candidates into true positives and false positives in a binary classification setting. We evaluate the classification accuracy by Area Under the Curve (AUC) score at candidate-level. 

\medskip
\noindent\textbf{LIDC-IDRI:} 
The Lung Image Database Consortium image collection (LIDC-IDRI) dataset is created by seven academic centers and eight medical imaging companies. The dataset includes 1,018 chest CT scans and marked-up annotated lung nodules.  The dataset is divided into training (510), validation (100), and test (408) sets. We pre-processed the data by re-sampling the 3D volumes to 1-1-1 spacing and then extracting a 64$\times$64$\times$32 crop around each nodule. The models are trained to segment long nodules in these 3D crops. The segmentation accuracy is measured by the Intersection over Union (IoU) metric.

\medskip
\noindent\textbf{LiTS:} The dataset is provided by MICCAI 2017 LiTS Challenge, including 130 CT scans with expert ground-truth segmentation masks for liver and tumor lesions.  We divide dataset into training (100 patients), validation (15 patients), and test (15 patients) sets. The models are trained to segment liver in 3D scans. The segmentation accuracy is measured by the Intersection over Union (IoU) metric.

\medskip
\noindent\textbf{BraTS:} The dataset includes brain MRI scans of 285 patients (210 HGG and 75 LGG) and segmentation ground truth for necrotic and non-enhancing tumor core, peritumoral
edema, GD-enhancing tumor, and background. For each patient, four different MR volumes are available: native
T1-weighted (T1), post-contrast T1-weighted (T1Gd), T2-weighted (T2), and T2 fluid attenuated inversion recovery (FLAIR). 
We divide dataset at patient-level into training (190 patients) and testing (95 patients) sets. The models are trained to segment brain tumors (background as negatives class and tumor sub-regions as positive class). The segmentation accuracy is measured by the Intersection over Union (IoU) metric.

\section{Implementation}
\subsection{Pre-training settings}
We apply DiRA to four existing self-supervised methods~\cite{chen2020improved, Chen2021Exploring, zbontar2021barlow, haghighi2021transferable}. To be self-contained, we'll explain each method briefly here. Also, we provide additional pre-training details that supplements~\cref{sec:pretraining_setup}.

\noindent\textbf{MoCo-v2~\cite{chen2020improved}:} We adopt MoCo-v2--- a popular representative of \textit{contrastive learning} methods, into our framework. MoCo leverages a momentum encoder to ensure the consistency of negative samples as they evolve during training. Moreover, a queue $K=\{k_1,k_2,...k_N\}$ is utilized to store the representations of negative samples. The discrimination task is to contrast  representations of positive and negative samples. As MoCo-v2 is adopted in DiRA, the encoder $f_\theta$ and  projection head $h_\theta$ are updated by back-propagation, while  $f_\xi$ and $h_\xi$ are updated by using an exponential moving average (EMA) of the parameters in $f_\theta$ and $h_\theta$, respectively. The discrimination branch is trained using InfoNCE loss~\cite{oord2019representation}, which for a pair of positive samples $x_1$ and $x_2$ defined as follows:

\begin{equation}
  \mathcal{L}_{dis} = - log \frac{exp(z_1\cdot z_2/\tau)}{\sum\limits_{n=0}^{N} exp(z_1 \cdot k_n/\tau)}
  \label{eq:nce_loss}
\end{equation}
where $z_1=h_\theta(f_\theta(x_1))$ and $z_2=h_\xi(f_\xi(x_2))$, $\tau$ is a temperature hyperparameter, and $N$ is the queue size.
Following~\cite{chen2020improved}, $f_\theta$ is a standard ResNet-50 and $h_\theta$ is a two-layer MLP head (hidden layer 2048-d, with ReLU). Moreover, when adopting MoCo-v2 in DiRA, $f_\theta$, $h_\theta$, and  $g_\theta$ are optimized using SGD  with an initial learning rate of 0.03, weight decay 0.0001, and the SGD momentum 0.9. 

\medskip
\noindent\textbf{SimSiam~\cite{Chen2021Exploring}:} We adopt SimSiam--- a popular representative of \textit{asymmetric instance discrimination} methods, into our framework. SimSiam trains the model without negative pairs and directly maximizes the similarity of two views from an image using a simple siamese network followed by a predictor head. To prevent collapsing solutions, a stop-gradient operation is utilized. As such, the model parameters are only updated using one distorted version of the input, while the representations from another distorted version are used as a fixed target. As SimSiam is adopted in DiRA, the encoder $f_\theta$ and  projection head $h_\theta$ share weights with  $f_\xi$ and $h_\xi$, respectively.  The model is trained to maximize the agreement between the representations of positive samples using negative cosine similarity, defined as follows:

\begin{equation}
  \mathcal{D}(z_1,y_2) = - \frac{z_1}{{\| z_1 \|}_2}\cdot\frac{y_2}{{\| y_2 \|}_2}
  \label{eq:nce_loss}
\end{equation} 
where $z_1=h_\theta(f_\theta(x_1))$ and $y_2=f_\xi(x_2)$.  The discrimination branch is trained using a symmetrized loss as follows:
\begin{equation}
  \mathcal{L}_{dis} = \frac{1}{2}\mathcal{D}(z_1,stopgrad(y_2)) + \frac{1}{2}\mathcal{D}(z_2,stopgrad(y_1)) 
  \label{eq:nce_loss}
\end{equation} 
where stopgrad means that $y_2$ is treated as a constant in this term. Following~\cite{Chen2021Exploring}, $f_\theta$ is a standard ResNet-50 and $h_\theta$ is a three-layer projection MLP head (hidden layer 2048-d), followed by a two-layer predictor MLP head. Moreover, when adopting SimSiam in DiRA, $f_\theta$, $h_\theta$, and  $g_\theta$ are optimized using SGD  with a linear scaling learning rate (lr$\times$BatchSize/256). The initial learning rate is 0.05, weight decay is 0.0001, and the SGD momentum is 0.9. 

\medskip
\noindent\textbf{Barlow Twins~\cite{zbontar2021barlow}:} 
We adopt Barlow Twins--- a popular representative of \textit{redundancy reduction instance discrimination learning} methods, into our framework.
Barlow Twins makes the cross-correlation matrix computed from two siamese branches close to the identity matrix. By equating the diagonal elements of the cross-correlation matrix to 1, the representation will be invariant to the distortions applied to the samples. By equating the off-diagonal elements of the cross-correlation matrix to 0, the different vector components of the representation will be decorrelated, so that the output units contain non-redundant information about the sample. The discrimination loss is defined as follows:

\begin{equation}
  \mathcal{L}_{dis} = \sum\limits_{i} (1 - \mathcal{C}_{ii})^2 + \lambda \sum\limits_{i} \sum\limits_{i\neq j} \mathcal{C}_{ij}^2
  \label{eq:nce_loss}
\end{equation}

where $\mathcal{C}$ is the cross-correlation matrix computed between the outputs of the $h_\theta$ and $h_\xi$ networks along the batch dimension. $\lambda$ is a coefficient that determines the importance of the invariance term and redundancy reduction term in the loss.
Following~\cite{zbontar2021barlow}, $f_\theta$ is a standard ResNet-50 and $h_\theta$ is a three-layer MLP head. Moreover, when adopting Barlow Twins in DiRA, $f_\theta$, $h_\theta$, and  $g_\theta$ are optimized using LARS optimizer  with the learning rate schedule similar to ~\cite{zbontar2021barlow}. 

\medskip
\noindent\textbf{TransVW~\cite{haghighi2021transferable}:} TransVW defines the similar anatomical patterns within medical images as anatomical visual words, and combines the discrimination and restoration of visual words in a single loss objective. As TransVW is adopted in DiRA, the encoder $f_\theta$ and  projection head $h_\theta$ are identical to $f_\xi$ and $h_\xi$, respectively.
In particular, the discrimination branch is trained to classify instances of visual words according to their pseudo class labels using the standard cross-entropy loss:  
\begin{equation}\label{eq:cls_loss}
\mathcal{L}_{dis}=-\frac{1}{B}\sum_{b=1}^{B}\sum_{c=1}^{C}\mathcal{Y}_{bc}\log  {\mathcal{P}_{bc}}
\end{equation}
where $B$ denotes the batch size; $C$ denotes the number of visual words classes; $\mathcal{Y}$ and $\mathcal{P}$ represent the ground truth (one-hot pseudo label vector obtained from visual word classes) and the prediction of $h_\theta$, respectively. Following~\cite{haghighi2021transferable}, we use 3D U-Net as the $f_\theta$ and $g_\theta$.  $h_\theta$ includes a set of fully-connected layers followed by a classification head. $f_\theta$ and $g_\theta$ are trained with the same setting as ~\cite{haghighi2021transferable}.

\medskip
\noindent\textbf{Joint training process:} Following~\cite{Chaitanya2020Contrastive,Chen2021Joint}, we perform the overall pre-training with the discrimination, restoration, and adversarial losses in a gradual evolutionary manner. First, the encoder $f_\theta$ along with projector $h_\theta$ are optimized using the discrimination loss $\mathcal{L}_{dis}$ according to the learning schedule of the original discriminative methods~\cite{chen2020improved,Chen2021Exploring,zbontar2021barlow,haghighi2021transferable}, empowering the model with an initial discrimination ability. Then, the restoration and adversarial losses are further fused into the training process incrementally. To stabilize the adversarial training process and reduce the noise from imperfect restoration at initial epochs~\cite{Chen2021Joint}, we first warm up the $f_\theta$ and $g_\theta$ using the $\mathcal{L}_{dis}+ \mathcal{L}_{res}$, and then add the adversarial loss $\mathcal{L}_{adv}$ to jointly train the whole framework; the optimization of the framework by incorporation of  $\mathcal{L}_{res}$ and $\mathcal{L}_{adv}$ takes up to 800 epochs. Following ~\cite{Zhou2021Preservational}, we use the early-stop technique on the validation set, and the checkpoints with the lowest validation loss are used for fine-tuning. 

\subsection{Fine-tuning settings}
\noindent\textbf{Preprocessing and data augmentation:} 
Following~\cite{Taher2021Systematic}, for 2D target tasks on X-ray datasets (ChestX-ray14, CheXpert, SIIM-ACR, and Montgomery), we resize the images to 224$\times$224. For thorax diseases classification tasks on ChestX-ray14 and CheXpert, we apply standard data augmentation techniques, including random cropping and resizing, horizontal flipping, and rotating. For segmentation tasks on SIIM-ACR and Montgomery, we apply  random brightness contrast, random gamma, optical distortion, elastic transformation, and grid distortion. For 3D target tasks, we use regular  data  augmentations  including  random  flipping,  transposing,  rotating, and adding Gaussian noise.

\noindent\textbf{Training parameters:}
We endeavour to optimize each downstream task with the best performing hyper-parameters. In all 2D and 3D downstream tasks, we use Adam optimizer with $\beta_1= 0.9$, $\beta_2= 0.999$. We use early-stop mechanism using the 10\% of the training data as the validation set to avoid over-fitting. For 2D classification tasks on ChestX-ray14 and CheXpert datasets, we use a learning rate $2e-4$ and \emph{ReduceLROnPlateau} as the learning rate decay scheduler. For 2D segmentation tasks on SIIM-ACR and Montgomery, we use a learning rate $1e-3$ and \emph{cosine} learning rate decay scheduler. For all 3D downstream tasks, we use \emph{ReduceLROnPlateau} as the learning rate decay scheduler. For downstream tasks on LUNA, PE-CAD, LIDC, and LiTS, we use a learning rate $1e-2$. For BraTS dataset, we use a learning rate of $1e-3$.

\end{document}